\documentclass[letterpaper]{article} 
\usepackage{aaai2026}  
\usepackage{times}  
\usepackage{helvet}  
\usepackage{courier}  
\usepackage[hyphens]{url}  
\usepackage{graphicx} 
\usepackage{natbib}  
\usepackage{caption} 
\usepackage{amsmath,amssymb}
\usepackage{amsthm}
\usepackage{booktabs} 
\usepackage{tikz}
\usepackage{ctable} 
\usepackage{subcaption}
\usepackage{multirow}
\usepackage[table]{xcolor}
\definecolor{headerblue}{RGB}{52, 205, 249}   
\definecolor{cellblue}{RGB}{203, 206, 251}    
\usepackage{algorithm}
\usepackage[noend]{algpseudocode}
\usepackage{xcolor}
\algrenewcommand\algorithmiccomment[1]{\hfill\textcolor{gray}{\(\triangleright\) #1}}
\newtheorem{prop}{Proposition}
\newtheorem{definition}{Definition}

\usepackage{enumitem}

\renewcommand{\arraystretch}{1}            
\newcommand{\methodpanel}[2]{%
  \subcaptionbox{\scriptsize #2}[0.153\textwidth]{%
    \begin{minipage}{\linewidth}
      \includegraphics[width=\linewidth,trim={1.1cm 0 0 1.1cm},clip]{#1}%
      \vspace{6pt} 
    \end{minipage}}%
}
\usepackage{tikz}
\usepackage{pgfplots}
\usepackage{subcaption}
\pgfplotsset{compat=1.18}
\usetikzlibrary{arrows.meta,positioning,fit,calc,shapes,decorations.pathmorphing}

\tikzset{
  box/.style={rounded corners, draw, thick, inner sep=3pt, fill=gray!5},
  arrow/.style={-{Latex[length=3mm]}, thick},
  emph/.style={draw=black, very thick},
}

\usepackage{newfloat}
\usepackage{listings}
\DeclareCaptionStyle{ruled}{labelfont=normalfont,labelsep=colon,strut=off} 
\floatstyle{ruled}
\newfloat{listing}{tb}{lst}{}
\floatname{listing}{Listing}
\nocopyright

\title{Multi-Objective Multi-Agent Path Finding with Lexicographic Cost Preferences}
\author {
    Pulkit Rustagi\textsuperscript{\rm 1},
    Kyle Hollins Wray\textsuperscript{\rm 2},
    Sandhya Saisubramanian\textsuperscript{\rm 1}
}
\affiliations {
    \textsuperscript{\rm 1}Oregon State University, Corvallis, OR, USA\\
    \textsuperscript{\rm 2}Northeastern University, Boston, MA, USA\\
    rustagip@oregonstate.edu, k.wray@northeastern.edu, sandhya.sai@oregonstate.edu
}

\begin{document}

\maketitle

\begin{abstract}
Many real-world scenarios require multiple agents to coordinate in shared environments, while balancing trade-offs between multiple, potentially competing objectives. Current multi-objective multi-agent path finding (MO-MAPF) algorithms typically produce conflict-free plans by computing Pareto frontiers. They do not explicitly optimize for user-defined preferences, even when the preferences are available, and scale poorly with the number of objectives. We propose a lexicographic framework for modeling MO-MAPF, along with an algorithm \textit{Lexicographic Conflict-Based Search} (LCBS) that directly computes a single solution aligned with a lexicographic preference over objectives. LCBS integrates a priority-aware low-level $A^*$ search with conflict-based search, avoiding Pareto frontier construction and enabling efficient planning guided by preference over objectives. We provide insights into optimality and scalability, and empirically demonstrate that LCBS computes optimal solutions while scaling to instances with up to ten objectives---far beyond the limits of existing MO-MAPF methods. Evaluations on standard and randomized MAPF benchmarks show consistently higher success rates against state-of-the-art baselines, especially with increasing number of objectives.
\end{abstract}

\begin{links}
    \link{Code}{https://github.com/Intelligent-Reliable-Autonomous-Systems/LCBS}
\end{links}

\section{Introduction}
  
Multi-Agent Path Finding (MAPF) addresses the problem of computing collision-free paths for a team of agents operating in a shared environment~\cite{standley2010finding}. MAPF enables coordinated planning in domains such as robotic warehouses~\cite{wurman2008coordinating}, airport ground operations~\cite{morris2016planning}, and multi-vehicle transport networks~\cite{chen2024traffic}. Multi-objective MAPF (MO-MAPF) supports MAPF settings with multiple objectives~\cite{ren2021multi}. This paper focuses on MO-MAPF settings where an inherent preference ordering over objectives is available.

For example, agents in a robotic warehouse must coordinate to minimize delivery time, avoid human worker zones, and minimize energy consumption. The objectives often follow a clear priority: human safety $\succ$ delivery time $\succ$ energy consumption. In such settings, leveraging the available preference ordering can improve the efficiency and scalability of the solution approach.

\begin{figure}[!t]
    \centering
    \includegraphics[width=\linewidth]{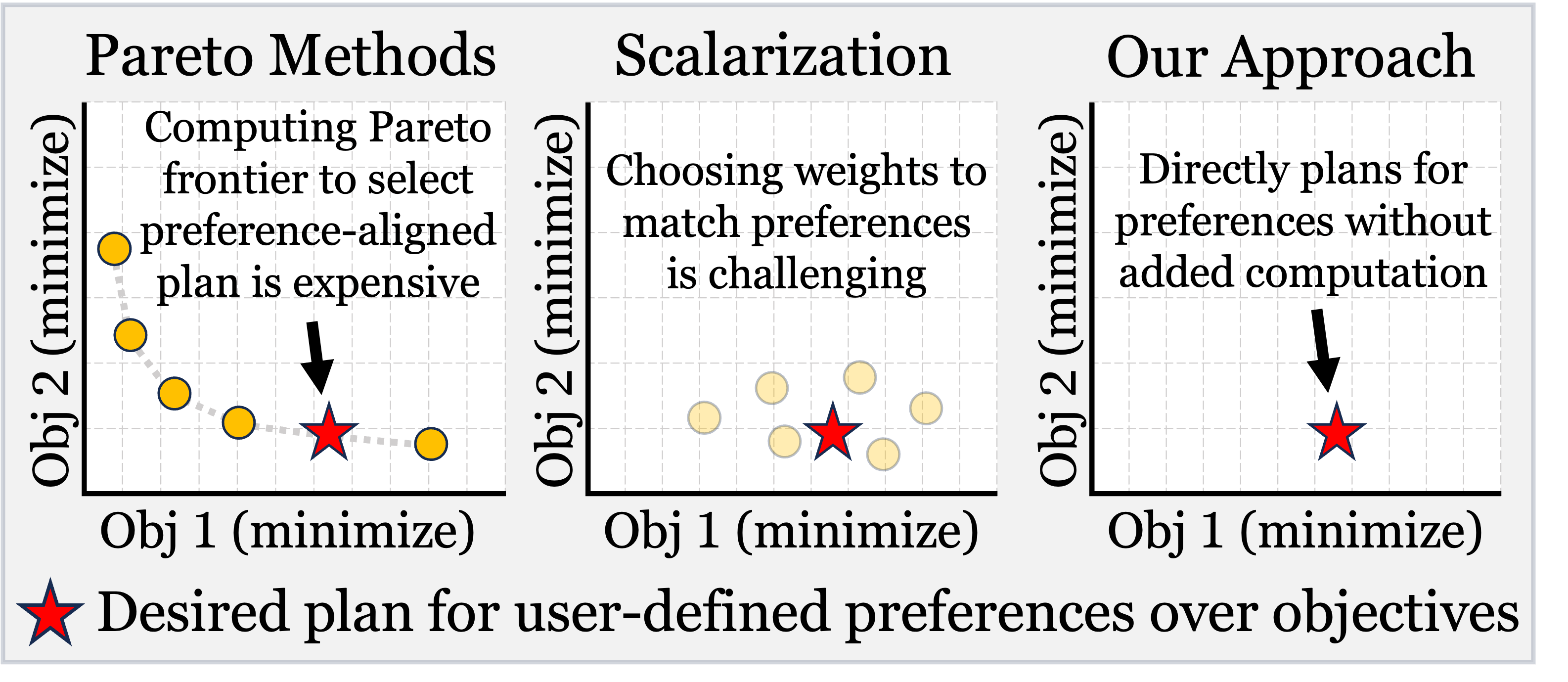}
    \caption{An illustration of multi-objective planning with user-defined preferences over two objectives. Unlike existing methods, our proposed approach can directly optimize objective preferences and does not require computing a Pareto frontier or determining weights for scalarization.}
    \label{fig:scope_of_this_work}
\end{figure}

Existing MO-MAPF methods typically construct a Pareto front of non-dominated joint plans~\cite{wang2024efficient,ren2021multi,ren2023binary}, often using frontier expansion~\cite{weise2020scalable} or evolutionary methods~\cite{ren2021subdimensional}. Even when a preference order over objectives is specified, these methods compute the Pareto front \emph{before} a preference-aligned solution is selected from it, making them computationally expensive and often intractable as the number of objectives increases. While scalarization-based methods~\cite{ho2023preference} reduce multiple objectives to a single cost, choosing weights to match preference is a non-trivial problem~\cite{wray2015multi}. Figure~\ref{fig:scope_of_this_work} summarizes limitations of current approaches in planning for user-defined preferences.

\begin{figure*}[t]
    \centering
    \includegraphics[width=\linewidth,trim={0 0.35cm 0 0},clip]{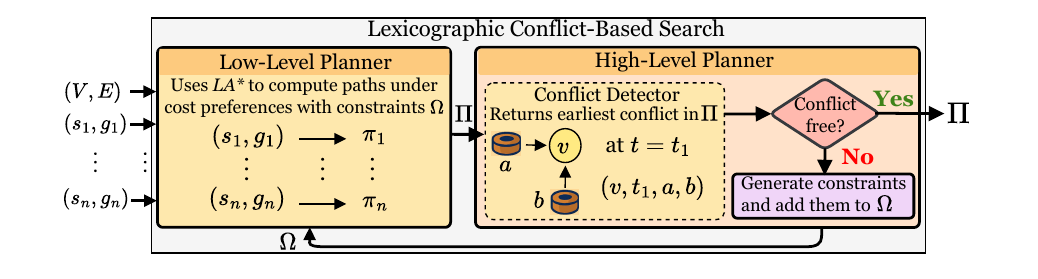}
    \caption{LCBS Algorithm Overview. The algorithm starts at the \emph{Low-Level Planner}, and uses lexicographic A$^*$ (LA$^*$) (Algorithm~\ref{alg:LA*}) to compute  shortest agent-paths to goal following lexicographic ordering over costs, without violating any constraints in $\Omega$ (initially, $\Omega=\emptyset$). At the high level (dotted), LCBS compiles the joint plan $\Pi$ and uses the \emph{Conflict Detector} module to get information about the conflict (\emph{location}, \emph{time}, \emph{agent 1}, \emph{agent 2}) if one exists, for example, agents \emph{a} and \emph{b} in conflict at vertex $v$ (or edge $e$) at time $t=t_1$ is encoded as $(v,t_1,a,b)$. If $\Pi$ is conflict-free, it is returned, else constraints are imposed on each agent, generated and evaluated using binary branching constraint tree~\cite{ren2023binary}. The constraint set $\Omega$ is updated and returned to the low-level planner that recomputes all agents paths. This process is iteratively done till a conflict-free $\Pi$ is found.}
    \label{fig:LCBS-overview}
\end{figure*}

We propose a \emph{lexicographic MO-MAPF} framework to efficiently model problems with lexicographic ordering over cost functions associated with each objective. We introduce an algorithm, \emph{Lexicographic Conflict-Based Search} (LCBS), to solve lexicographic MAPF by directly computing a single solution aligned with the objective ordering.
LCBS builds on Conflict-Based Search (CBS) framework~\cite{sharon2015conflict} by replacing the low-level search with a lexicographic $A^*$ planner that sequentially optimizes each objective in decreasing order of priority. At the high level, LCBS resolves conflicts using constraint branching~\cite{wang2024efficient}, preserving cost structure and enabling scalability. 

Our evaluation on standard and randomized MO-MAPF benchmarks shows that LCBS achieves consistently higher success rates than state-of-the-art baselines. Evaluation results with up to ten objectives and 35 agents demonstrate the scalability of our approach.

\section{Multi-Objective MAPF}

Multi-Objective MAPF (MO-MAPF) extends classical MAPF to settings with multiple objectives, each associated with a cost function. Agents $A = \{1, \dots, n\}$ move in discrete time on a graph $G = (V, E)$, with each agent $i \in A$ following a path $\pi_i$ from start $s_i$ to goal $g_i$. A joint plan $\Pi = (\pi_1, \dots, \pi_n)$ is \emph{valid} if it avoids vertex and edge conflicts~\cite{stern2019multi}. Each agent incurs a cost vector $\mathbf{c}_i \in \mathbb{R}_+^d$, yielding total cost $\mathbf{C}(\Pi) = \sum_{i \in A} \mathbf{c}_i$. 
Classical MAPF is a special case with $d = 1$, where costs are scalar and preferences are not modeled. Solving MO-MAPF typically involves generating non-dominated joint plans along the Pareto frontier, returning solutions that balance trade-offs across objectives. However, such methods do not explicitly encode objective preferences, often available from the environment or task context. Instead, preferences are applied only \emph{after} solution generation, limiting their influence on the planning process.

\paragraph{Lexicographic MO-MAPF}  
In many settings, user-defined preferences induce a clear priority ordering over cost objectives. Lexicographic MO-MAPF formalizes this by optimizing cost vectors under a strict ordering $C^1\!\succ\!\dots \succ C^d$, where $C^1$ and $C^d$ denote the highest and lowest priority cost dimensions of $\mathbf{C}(\Pi) = (C^1, \dots, C^d)$. This enables preference-aware planning by enforcing priority directly during optimization. Such prioritization is often natural and available in practical applications, e.g., prioritizing safety over energy in human-robot interaction. A lexicographic ordering ensures that higher-priority objectives are optimized before considering lower-priority ones.

\begin{definition}[Lexicographic comparison $<_{\text{lex}}$]
\label{def:lex}
For $\mathbf{c}_1,\mathbf{c}_2\in\mathbb{R}_+^d$, let 
$j=\min\{k\in\{1,\dots,d\}\mid c_1^k\neq c_2^k\}$.
We write $\mathbf{c}_1<_{\text{lex}}\mathbf{c}_2$ iff
$c_1^k=c_2^k$ for all $k<j$ and $c_1^j<c_2^j$.
\end{definition}

\paragraph{Optimality Criteria}
A $\bar{\mathbf{\Pi}}$ be the set of all valid joint plans. The joint plan $\Pi^\star\in\bar{\mathbf{\Pi}}$ is optimal if its total cost vector is lexicographically minimum. Formally, 
\begin{equation}
\label{eq:lex-optimality}
\mathbf{C}(\Pi^\star)<_{\text{lex}}\mathbf{C}(\Pi'), \quad\forall \,\Pi'\in \bar{\mathbf{\Pi}}\setminus\Pi^\star
\end{equation}
Equation~\ref{eq:lex-optimality} ensures that $\Pi^\star$ lexicographically minimizes costs from highest to lowest priority following Definition~\ref{def:lex}.

Below, we present an algorithm that builds on conflict-based search to solve lexicographic MO-MAPF.

\section{Lexicographic Conflict-Based Search}

LCBS extends the two-level framework of Conflict-Based Search (CBS)~\cite{sharon2012conflict} to settings with lexicographic preferences over objectives (Figure~\ref{fig:LCBS-overview}). As in CBS, the high-level planner maintains a constraint tree (CT)~\cite{sharon2015conflict}, where each node represents a joint plan and a set of constraints $\Omega$ that restrict agent movements over vertices or edges at specific timesteps. Upon detecting a conflict, the node is split into two children, each imposing a constraint on one of the conflicting agents. This process continues until a conflict-free plan is found. The low-level planner in LCBS is a \emph{lexicographic A$^*$} (LA$^*$) (Algorithm~\ref{alg:LA*}), which computes individual agent paths that sequentially optimize objectives from highest to lowest priority. We adopt the high-level planning structure from BB-MO-CBS-pex~\cite{wang2024efficient}, for resolving conflicts and merging joint paths.

\paragraph{Low-Level Search}
LCBS uses LA$^*$ to compute agent paths following the lexicographic order over costs. The search is done on time-augmented states $(v,t)$ with cumulative vector cost $\mathbf{g}(v,t)$ and admissible heuristic $\mathbf{h}(v)$. The key is $\mathbf{f}(v,t)=\mathbf{g}(v,t)+\mathbf{h}(v)$. States with lexicographically smaller $\mathbf{f}$ (under $<_{\text{lex}}$) are expanded first, ensuring higher-priority objectives are optimized before lower-priority ones.

Algorithm~\ref{alg:LA*} initializes the open list $\mathcal{O}$ with $(s,0)$ and maintains a closed map $\mathcal{C}$ to store the best $\mathbf{g}$-cost per state. In each iteration, LA$^*$ pops the state with lexicographically minimum $\mathbf{f}$ (Line 8). If the vertex is goal $g$ and waiting at $g$ is safe under $\Omega$, the path is returned (Lines 9-10). Else, the search expands moves to neighbors and a wait action, generating $(u,t{+}1)$ and skipping any transition that violates $\Omega$ (Lines 11-14). A successor is pushed (or re-opened) when its $\mathbf{g}$ improves under $<_{\text{lex}}$ (Lines 15-18). This continues until a feasible path is found or open list  $\mathcal{O}$ is exhausted.

\begin{algorithm}[t]
\caption{LA$^*$ (Lexicographic A$^*$)}
\label{alg:LA*}
\textbf{Input}: start $s$, goal $g$, graph $G=(V,E)$; edge cost $\mathbf{c}_e:E\!\to\!\mathbb{R}_+^d$ (wait included); heuristic $\mathbf{h}:V\!\to\!\mathbb{R}_+^d$ (admissible \& consistent per dimension); constraints $\Omega$\\
\textbf{Output}: Lexicographically optimal path $\pi$ from $s$ to $g$ (or $\emptyset$)
\begin{algorithmic}[1]
\State \textbf{States:} time-augmented $x=(v,t)$ with $v\in V$, $t\in\mathbb{N}$; 
\State \textbf{Neighbors:} $\mathcal{N}(v)\cup\{v\}$ (wait)
\State $x_0 \gets (s,0)$;\ $\mathbf{g}(x_0)\gets \mathbf{0}$;\ $\mathbf{f}(x_0)\gets \mathbf{h}(s)$
\State \textbf{Init:} open list $\mathcal{O}$ (priority by $<_{\text{lex}}$ on $\mathbf{f}$)
\State \textbf{Init:} closed map $\mathcal{C}$ from states to best $\mathbf{g}$ (lex order)
\State $\mathcal{O}.\textsc{Push}(x_0)$; \ $\mathcal{C}[x_0]\gets \mathbf{0}$
\While{$\mathcal{O}\neq\emptyset$}
  \State $x=(v,t) \gets \mathcal{O}.\textsc{PopMin}()$ \Comment{lex-min $\mathbf{f}$}
  \If{$v=g$ \textbf{and} \textbf{not }\textsc{Violates}$(x,\Omega)$}
     \State \Return \textsc{ReconstructPath}$(x)$
  \EndIf
  \For{each $u \in \mathcal{N}(v)\cup\{v\}$} \Comment{move or wait}
     \State $y \gets (u,t{+}1)$
     \If{\textsc{Violates}$((v,t)\!\to\!(u,t{+}1),\Omega)$} 
     \State \textbf{continue} \EndIf
     \State $\mathbf{g}' \gets \mathbf{g}(x) + \mathbf{c}_e(v,u)$;\ \ $\mathbf{f}' \gets \mathbf{g}' + \mathbf{h}(u)$
     \If{$y \notin \mathcal{C}$ \textbf{or} $\mathbf{g}' <_{\text{lex}} \mathcal{C}[y]$}
        \State $\mathcal{C}[y]\gets \mathbf{g}'$;\ \ \textsc{Parent}$(y)\gets x$;\ \ $\mathbf{f}(y)\gets \mathbf{f}'$
        \State $\mathcal{O}.\textsc{Push}(y)$
     \EndIf
  \EndFor
\EndWhile
\State \Return $\emptyset$
\end{algorithmic}
\end{algorithm}

\paragraph{High-Level Search}
LCBS adopts the high-level search from BB-MO-CBS-pex~\cite{wang2024efficient}, which maintains a constraint tree (CT) whose nodes store a joint plan ($\mathcal{N}.\Pi$), its joint cost vector ($\mathcal{N}.\mathbf{C}$), and the associated time-indexed constraint set ($\mathcal{N}.\Omega$). The high-level open list $\mathcal{O}_{\mathrm{HL}}$ is ordered lexicographically by $\mathcal{N}.\mathbf{C}$. We use standard helpers from BB-MO-CBS-pex~\cite{wang2024efficient}, to detect the earliest conflict and to generate the two child constraints (one per agent).

The root node $\mathcal{N}_0$ is built with $\Omega_0=\emptyset$ by employing LA$^*$ for each agent and computing $\mathcal{N}_0.\mathbf{C}=\sum_i \mathbf{c}(\pi_i)$ (Lines 1-7). The node is pushed to $\mathcal{O}_{\mathrm{HL}}$ (Line 8). At the start of each iteration, the loop pops the node with lexicographically smallest joint cost (Line 11). If no conflict is detected, the joint plan is returned (Lines 12-14). Otherwise, the earliest-time conflict is extracted and two children are created, each adding a single time-indexed vertex/edge constraint for one of the agents (Lines 15-16). The constrained agent is replanned with LA$^*$ (Lines 17-19) and feasible child node update their cost $\mathbf{C}$ before getting pushed into $\mathcal{O}_{\mathrm{HL}}$ (Lines 20-23). The process repeats until a conflict-free plan is obtained or the open list $\mathcal{O}_{\mathrm{HL}}$ is exhausted. 

With the open lists at both levels ordered by the lexicographic cost and component-wise consistent heuristics, the standard CBS correctness argument carries over. The first conflict-free node popped is lexicographically optimal. Vector comparisons introduce an $O(d)$ overhead per queue operation, and the CT structure remains unchanged.

\begin{algorithm}[t]
\caption{LCBS (High-Level Search)}
\label{alg:LCBS}
\textbf{Input}: $G=(V,E)$; agents $A=\{1,\dots,n\}$; start/goal pairs $(s^i,g^i)$; edge cost $\mathbf{c}_e$; heuristic $\mathbf{h}$; constraints over (vertex/edge, time)\\
\textbf{Output}: Conflict-free joint plan $\Pi=(\pi_1,\dots,\pi_n)$ (or $\emptyset$)
\begin{algorithmic}[1]
\State \textbf{Init:} root node $\mathcal{N}_0$ with $\Omega_0\!=\!\emptyset$
\For{each agent $i \in A$}
   \State $\pi_i \gets \textsc{LA*}(s^i,g^i,G,\mathbf{c}_e,\mathbf{h},\Omega_0)$
   \If{$\pi_i=\emptyset$} \State \Return $\emptyset$ \EndIf
\EndFor
\State $\mathcal{N}_0.\Pi \gets (\pi_1,\dots,\pi_n)$
\State $\mathcal{N}_0.\mathbf{C} \gets \sum_{i\in A}\mathbf{c}(\pi_i)$ \Comment{$\mathbf{c}(\pi_i)=\sum_{e\in \pi_i}\mathbf{c}_e(e)$}
\State \textbf{Init:} high-level open list $\mathcal{O}_{\mathrm{HL}}$ (priority by $<_{\text{lex}}$ on $\mathbf{C}$)
\State $\mathcal{O}_{\mathrm{HL}}.\textsc{Push}(\mathcal{N}_0)$
\While{$\mathcal{O}_{\mathrm{HL}}\neq\emptyset$}
   \State $\mathcal{N} \gets \mathcal{O}_{\mathrm{HL}}.\textsc{PopMin}()$ \Comment{lex-min joint cost}
   \State $\textit{conflict} \gets \textsc{DetectFirstConflict}(\mathcal{N}.\Pi)$ \Comment{earliest-time vertex/edge}
   \If{$\textit{conflict}=\varnothing$} \State \Return $\mathcal{N}.\Pi$ \EndIf
   \State $(\textit{type},\ell,t,i,j) \gets \textit{conflict}$ \Comment{conflict type, location, time, agents}
   \State $\{\omega^i,\omega^j\}\gets \textsc{GenerateConstraints}(\textit{conflict})$ 
   \For{each $a \in \{i,j\}$}
      \State Create child $\mathcal{N}_a$;\ \ $\mathcal{N}_a.\Omega \gets \mathcal{N}.\Omega \cup \{\omega^a\}$
      \State $\pi_a' \gets \textsc{LA*}(s^a,g^a,G,\mathbf{c}_e,\mathbf{h},\mathcal{N}_a.\Omega)$
      \If{$\pi_a' \neq \emptyset$}
         \State $\mathcal{N}_a.\Pi \gets$ $\mathcal{N}.\Pi$ with $\pi_a$ replaced by $\pi_a'$
         \State $\mathcal{N}_a.\mathbf{C} \gets \sum_{k\in A}\mathbf{c}(\pi_k)$
         \State $\mathcal{O}_{\mathrm{HL}}.\textsc{Push}(\mathcal{N}_a)$
      \EndIf
   \EndFor
\EndWhile
\State \Return $\emptyset$
\end{algorithmic}
\end{algorithm}

\setlength{\tabcolsep}{3pt}
\renewcommand{\arraystretch}{1}
\begin{figure*}[!t]
  \centering
  \includegraphics[width=\textwidth]{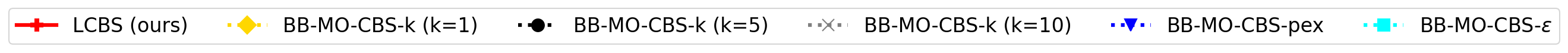}\vspace{3pt}

  \begin{tabular}{ccccc}
    \subcaptionbox{empty-32-32}[0.19\textwidth]{%
      \includegraphics[width=\linewidth]{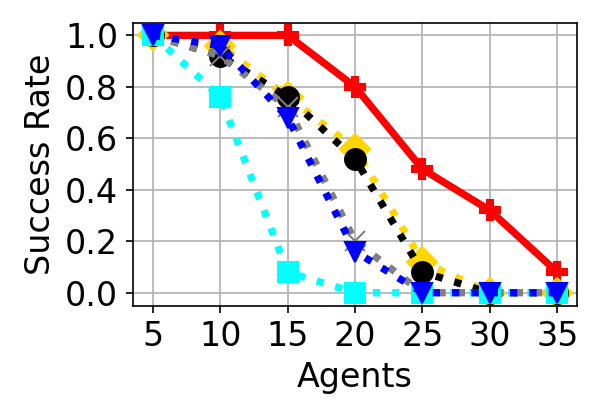}} &
    \subcaptionbox{empty-48-48}[0.19\textwidth]{%
      \includegraphics[width=\linewidth]{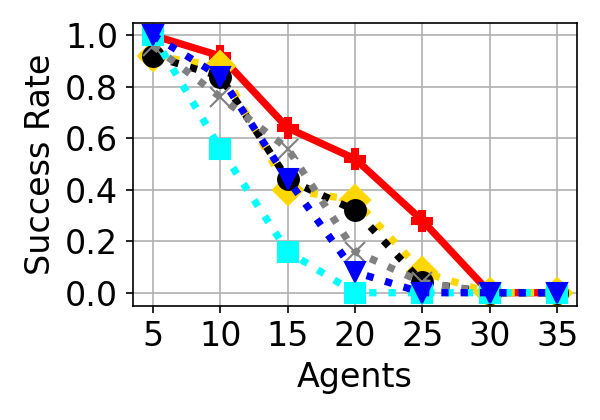}} &
    \subcaptionbox{maze-32-32-2}[0.19\textwidth]{%
      \includegraphics[width=\linewidth]{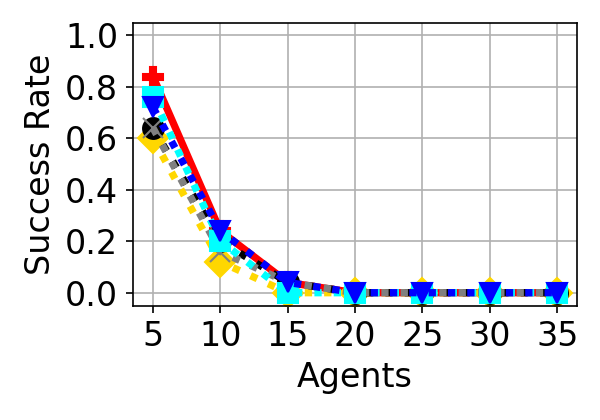}} &
    \subcaptionbox{maze-32-32-4}[0.19\textwidth]{%
      \includegraphics[width=\linewidth]{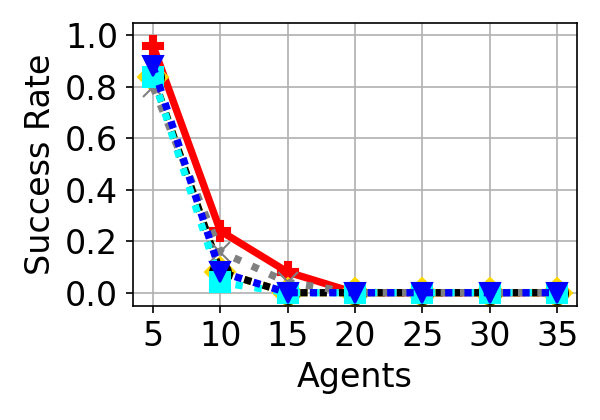}} &
    \subcaptionbox{random-32-32-20}[0.19\textwidth]{%
      \includegraphics[width=\linewidth]{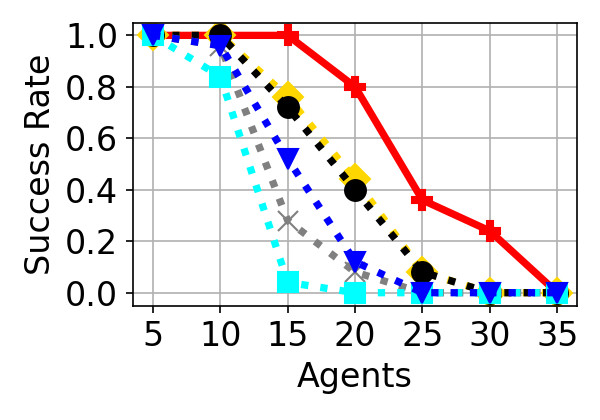}} \\
    \subcaptionbox{random-64-64-10}[0.19\textwidth]{%
      \includegraphics[width=\linewidth]{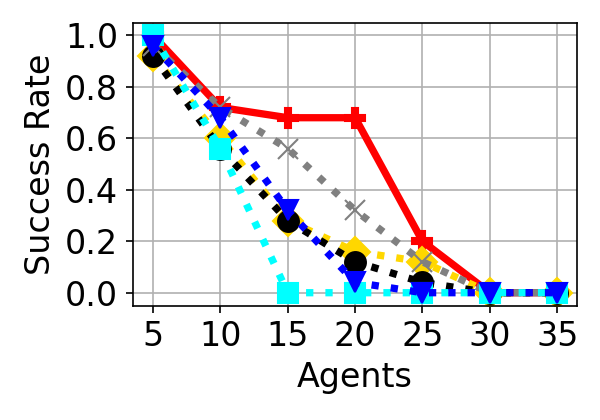}} &
    \subcaptionbox{random-64-64-20}[0.19\textwidth]{%
      \includegraphics[width=\linewidth]{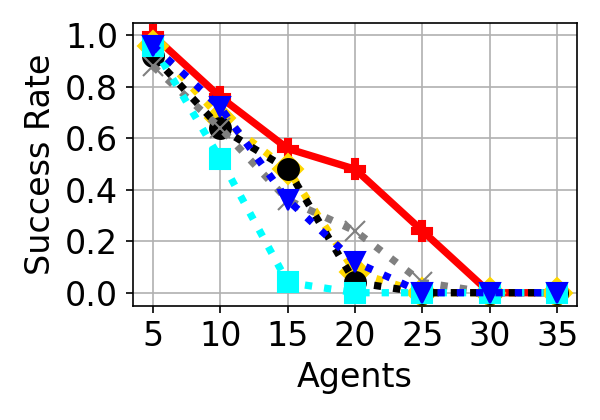}} &
    \subcaptionbox{room-32-32-4}[0.19\textwidth]{%
      \includegraphics[width=\linewidth]{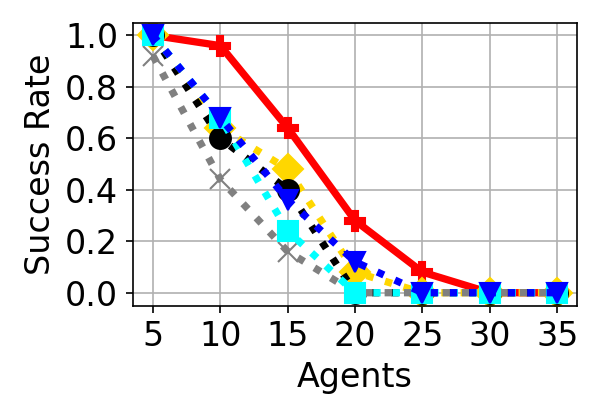}} &
    \subcaptionbox{room-64-64-8}[0.19\textwidth]{%
      \includegraphics[width=\linewidth]{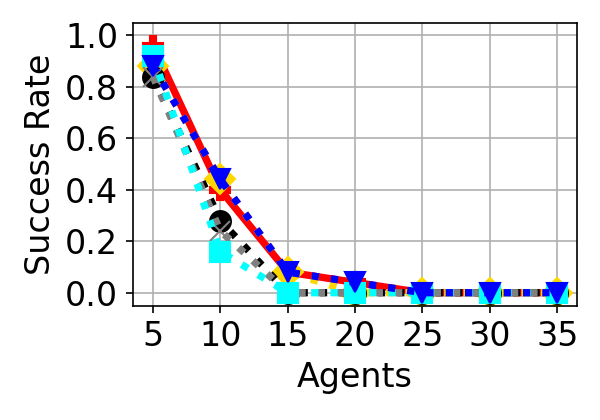}} &
    \subcaptionbox{warehouse-10-20-10-2-1}[0.19\textwidth]{%
      \includegraphics[width=\linewidth]{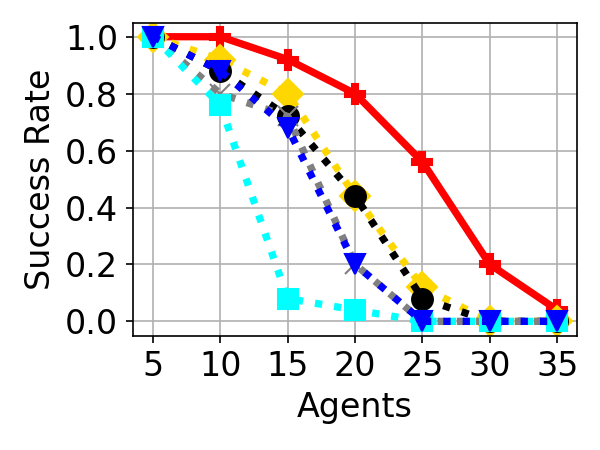}} \\
  \end{tabular}

  \caption{Success rate trends across MAPF benchmarks for 25 \emph{standard} scenarios with \textbf{3 objectives} $(T=2\text{ mins})$.}
  \label{fig:scenario-grid-3objs}
\end{figure*}

\begin{figure*}[t]
  \centering

  \begin{tabular}{ccccc}
    \subcaptionbox{empty-32-32}[0.19\textwidth]{\includegraphics[width=\linewidth]{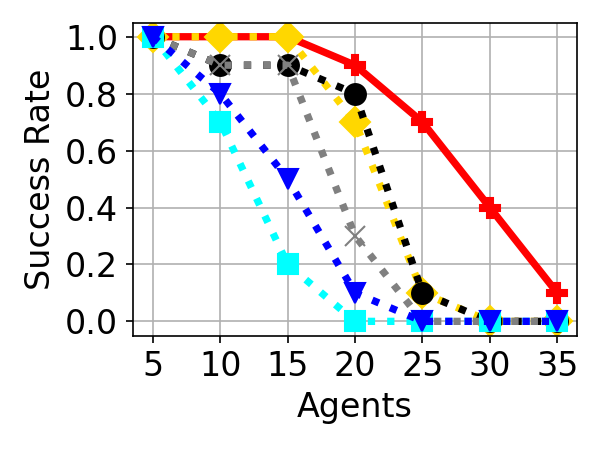}} &
    \subcaptionbox{empty-48-48}[0.19\textwidth]{\includegraphics[width=\linewidth]{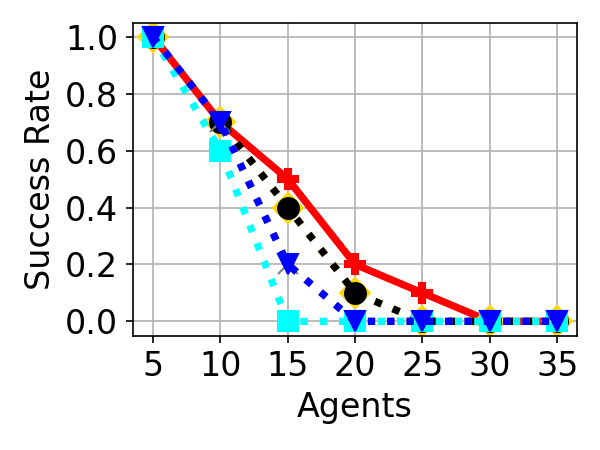}} &
    \subcaptionbox{maze-32-32-2}[0.19\textwidth]{\includegraphics[width=\linewidth]{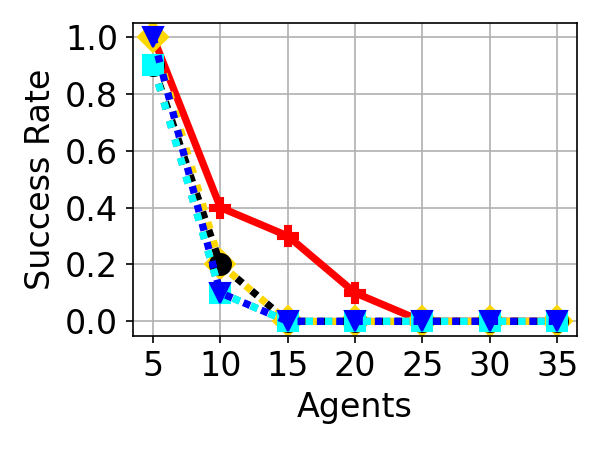}} &
    \subcaptionbox{maze-32-32-4}[0.19\textwidth]{\includegraphics[width=\linewidth]{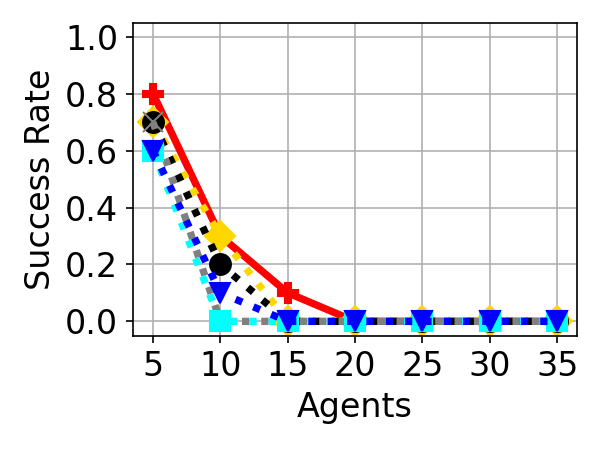}} &
    \subcaptionbox{random-32-32-20}[0.19\textwidth]{\includegraphics[width=\linewidth]{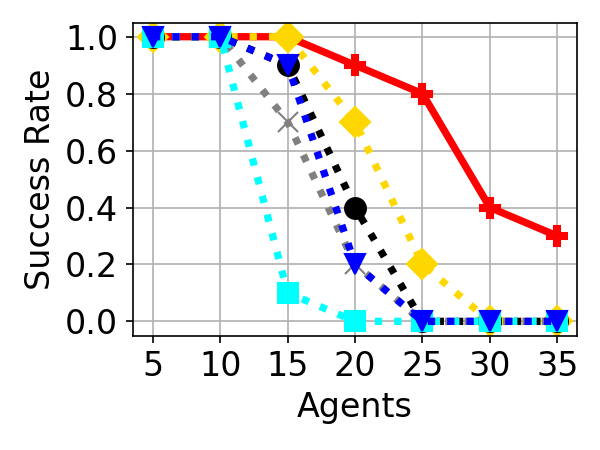}} \\
    \subcaptionbox{random-64-64-10}[0.19\textwidth]{\includegraphics[width=\linewidth]{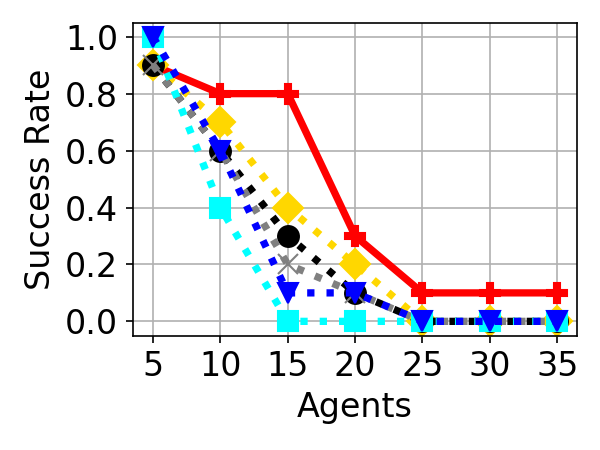}} &
    \subcaptionbox{random-64-64-20}[0.19\textwidth]{\includegraphics[width=\linewidth]{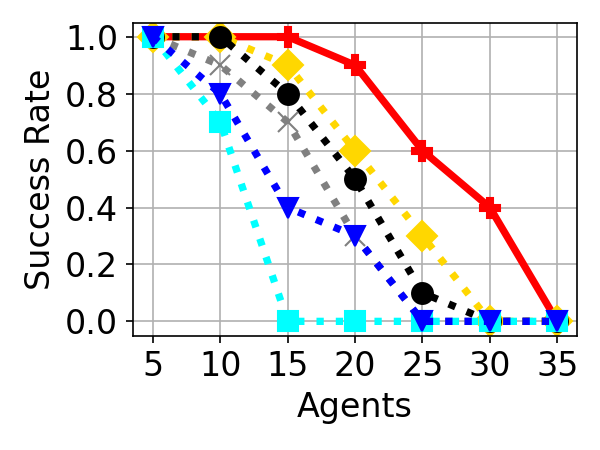}} &
    \subcaptionbox{room-32-32-4}[0.19\textwidth]{\includegraphics[width=\linewidth]{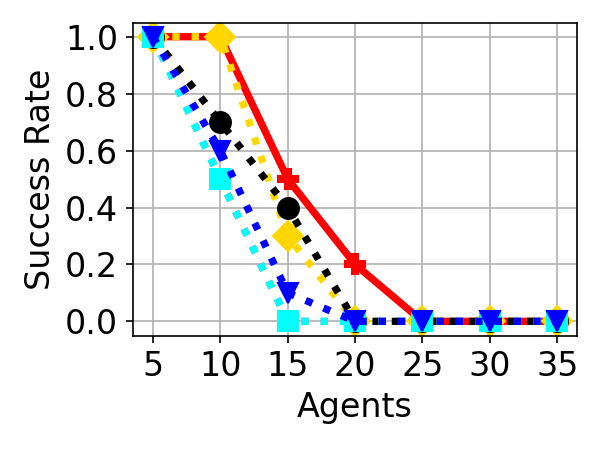}} &
    \subcaptionbox{room-64-64-8}[0.19\textwidth]{\includegraphics[width=\linewidth]{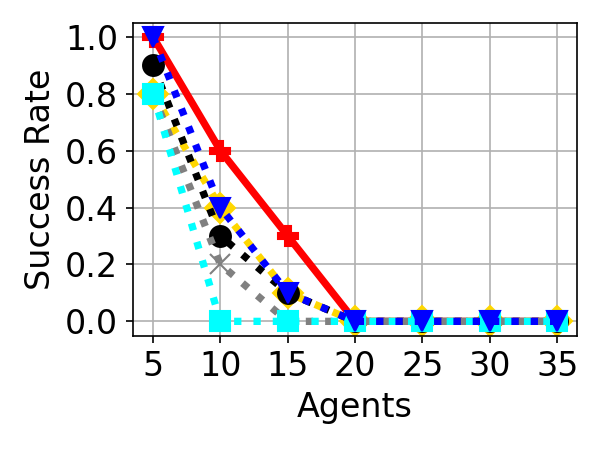}} &
    \subcaptionbox{warehouse-10-20-10-2-1}[0.19\textwidth]{\includegraphics[width=\linewidth]{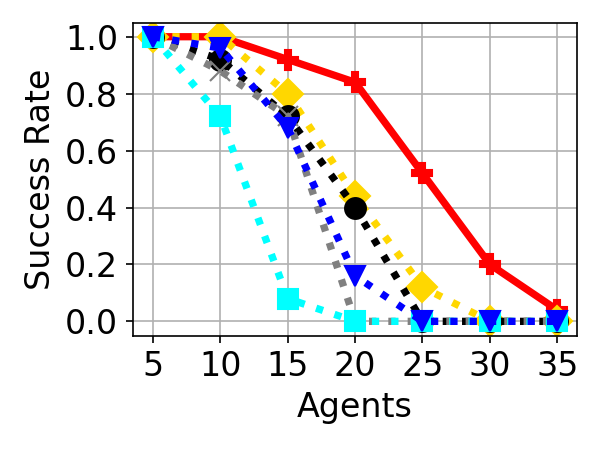}}
  \end{tabular}

  \caption{Success rate trends across MAPF benchmarks for 10 \emph{randomized} scenarios for \textbf{3 objectives} $(T=2\text{ mins})$.}
  \label{fig:scenario-grid-rand-3objs}
\end{figure*}


\paragraph{Additional Insights} 
We now provide two insights into LCBS's relation to the Pareto frontier and its scalability.
Importantly, the main contribution of this paper remains the LCBS algorithm and the suite of experiments in the next section.
These brief propositions below are included to provide context, and perhaps a direction for potential future theoretical work beyond the scope of this paper.

\begin{prop}\label{prop:lexicographic_is_optimal}
LCBS returns a solution on the Pareto front.
\end{prop}

\begin{proof}
We prove this by showing that LCBS returns a joint path that is lexicographically optimal and therefore a part of the Pareto frontier~\cite{junker2004preference}. 

Consider a high-level search node $N$ in the constraint tree with constraint set $\Omega$ (time-indexed vertex/edge constraints). In $N$, LA$^*$ orders the open list by $<_{\text{lex}}$ on $f=g+h$ (Alg. \ref{alg:LA*}, Line 4), pops the state with the lexicographic lowest $f$ (Line 8), and returns the first feasible path to goal (Lines 9–10). During expansion, successors that violate $\Omega$ are skipped (Lines 12–13). A successor is inserted/updated only when its $g$ cost improves under $<_{\text{lex}}$ as in Eq. \ref{eq:lex-optimality} (Lines 16–18). For fixed $\Omega$, the path chosen for each agent in $N$ is therefore lexicographically optimal with respect to Eq.~\ref{eq:lex-optimality}.


We now prove by contradiction that the joint plan returned by the high level (Algo.~\ref{alg:LCBS}) is lexicographically optimal under Eq.~\ref{eq:lex-optimality}. Let $\Pi$ be the conflict free plan returned by the high level and let $C(\Pi)=\sum_{i\in A} c(\pi_i)$. Assume $\Pi$ is not lexicographically optimal under Eq.~\ref{eq:lex-optimality}, so there exists a valid $\Pi'$ with $C(\Pi')<_{\text{lex}}C(\Pi)$. This implies that there exists at least one agent $i$ that lexicographically lowered its path cost, $c(\pi'_i)<_{\text{lex}}c(\pi_i)$. However, since we have already established 
that each agent path satisfies Eq.~\ref{eq:lex-optimality}, $C(\Pi)<_{\text{lex}}C(\Pi')$, this is a contradiction. Therefore $\Pi$, computed by Algo.~\ref{alg:LCBS}, is lexicographically optimal and lies on the Pareto frontier~\cite{junker2004preference}.  

\end{proof}

We now show that LCBS scales linearly in the number of objectives, a property that is particularly beneficial in high-dimensional multi-objective planning where Pareto-based  methods often struggle to find any solution.


\setlength{\tabcolsep}{3pt}
\renewcommand{\arraystretch}{1}
\begin{figure*}[!t]
  \centering
  \includegraphics[width=\textwidth]{figures/legend.png}\vspace{3pt}

  \begin{tabular}{ccccc}
    \subcaptionbox{empty-32-32}[0.19\textwidth]{%
      \includegraphics[width=\linewidth]{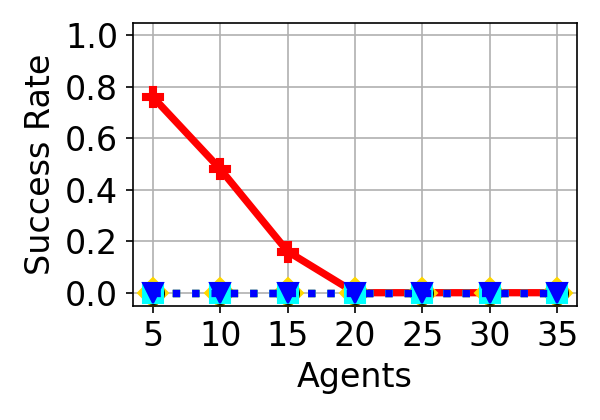}} &
    \subcaptionbox{empty-48-48}[0.19\textwidth]{%
      \includegraphics[width=\linewidth]{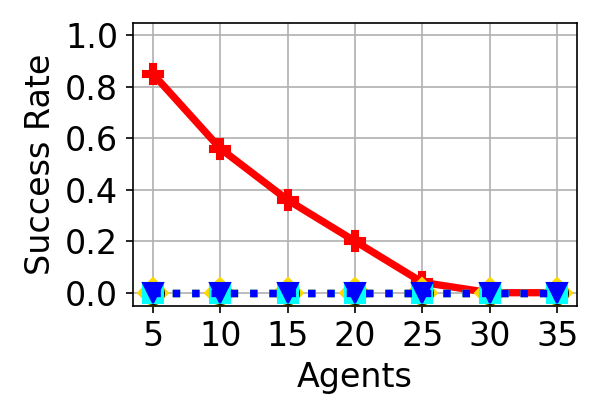}} &
    \subcaptionbox{maze-32-32-2}[0.19\textwidth]{%
      \includegraphics[width=\linewidth]{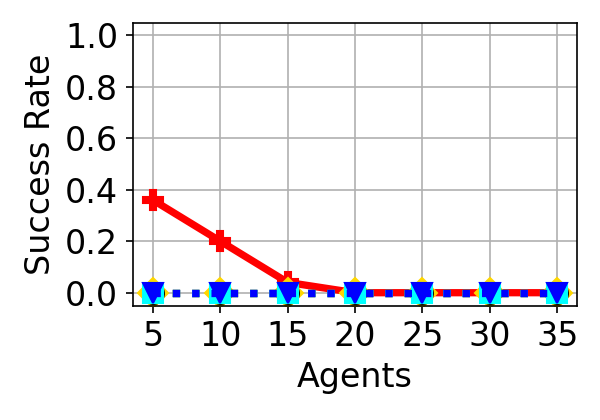}} &
    \subcaptionbox{maze-32-32-4}[0.19\textwidth]{%
      \includegraphics[width=\linewidth]{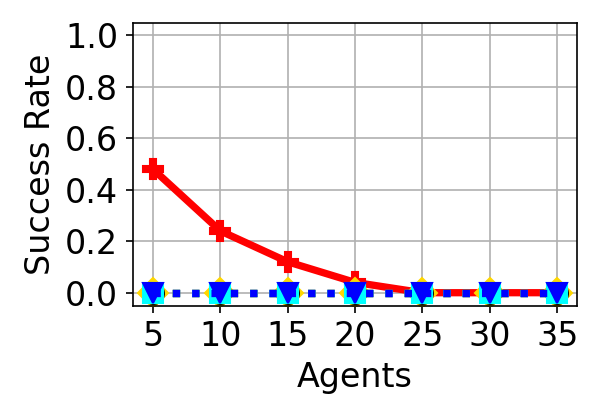}} &
    \subcaptionbox{random-32-32-20}[0.19\textwidth]{%
      \includegraphics[width=\linewidth]{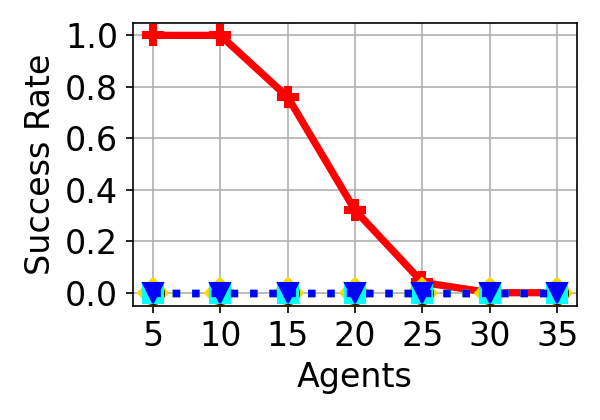}} \\
    \subcaptionbox{random-64-64-10}[0.19\textwidth]{%
      \includegraphics[width=\linewidth]{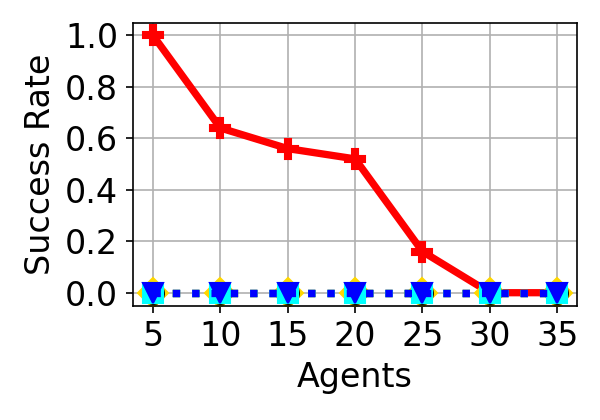}} &
    \subcaptionbox{random-64-64-20}[0.19\textwidth]{%
      \includegraphics[width=\linewidth]{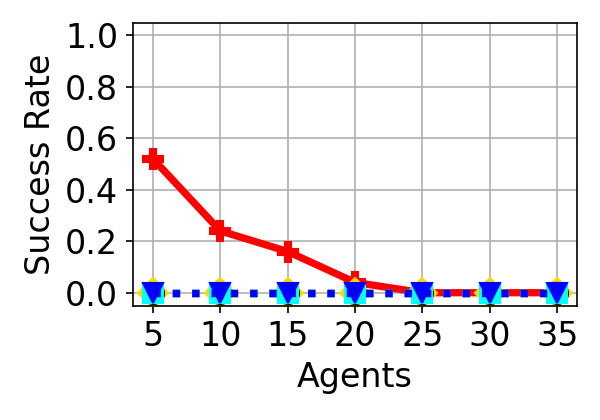}} &
    \subcaptionbox{room-32-32-4}[0.19\textwidth]{%
      \includegraphics[width=\linewidth]{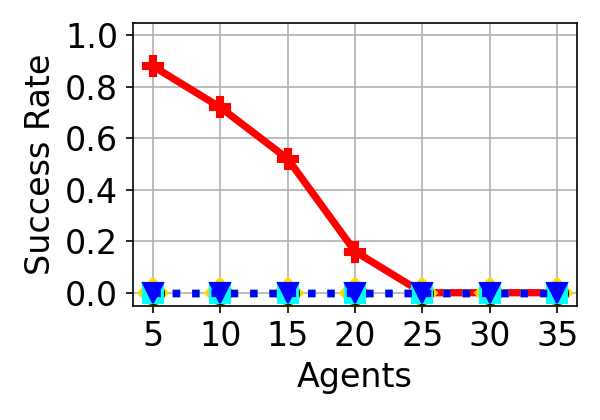}} &
    \subcaptionbox{room-64-64-8}[0.19\textwidth]{%
      \includegraphics[width=\linewidth]{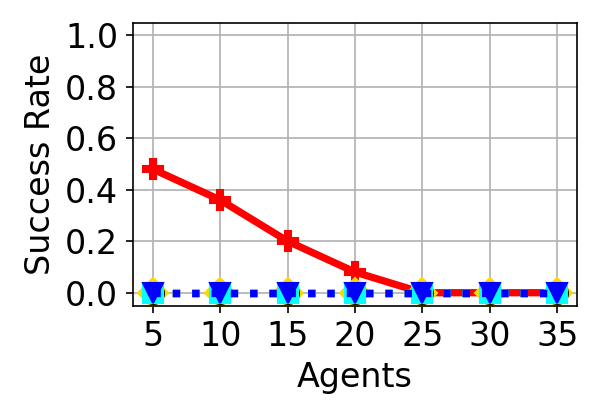}} &
    \subcaptionbox{warehouse-10-20-10-2-1}[0.19\textwidth]{%
      \includegraphics[width=\linewidth]{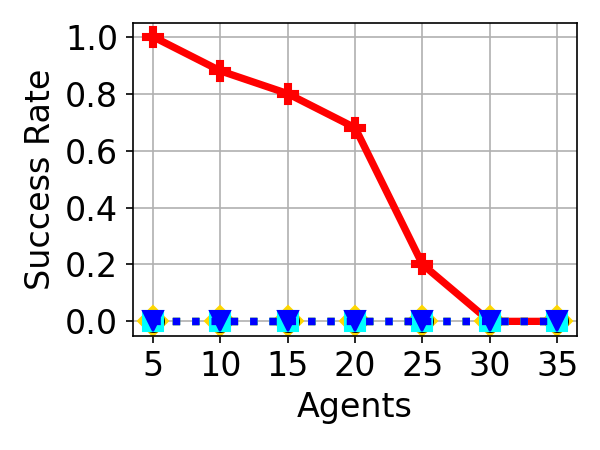}} \\
  \end{tabular}

  \caption{Success rate trends across MAPF benchmarks for 25 \emph{standard} scenarios with \textbf{4 objectives} $(T=2\text{ mins})$.}
  \label{fig:scenario-grid-4objs}
\end{figure*}


\begin{figure*}[!t]
  \centering

  \begin{tabular}{cccccccccc}
    \subcaptionbox{{empty-32-32}}[0.19\textwidth]{%
      \includegraphics[width=\linewidth]{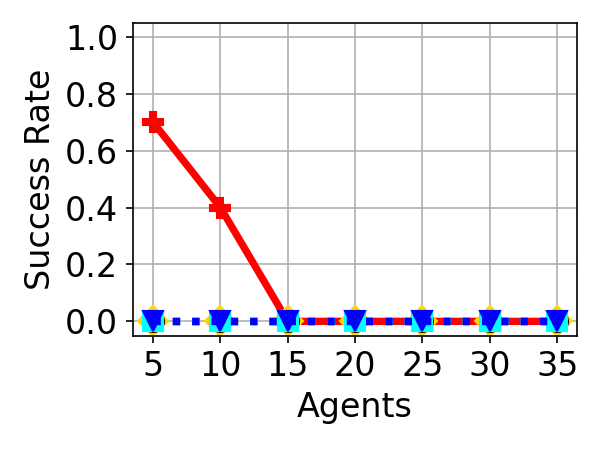}} &
    \subcaptionbox{{empty-48-48}}[0.19\textwidth]{%
      \includegraphics[width=\linewidth]{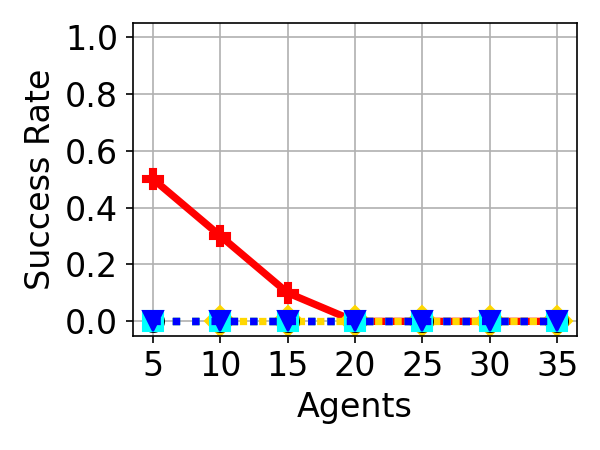}} &
    \subcaptionbox{{maze-32-32-2}}[0.19\textwidth]{%
      \includegraphics[width=\linewidth]{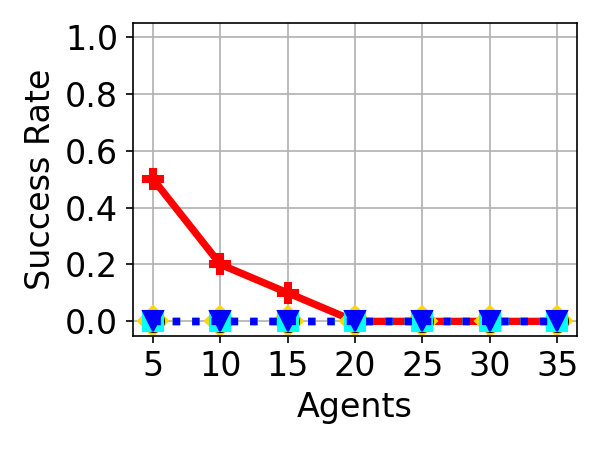}} &
    \subcaptionbox{{maze-32-32-4}}[0.19\textwidth]{%
      \includegraphics[width=\linewidth]{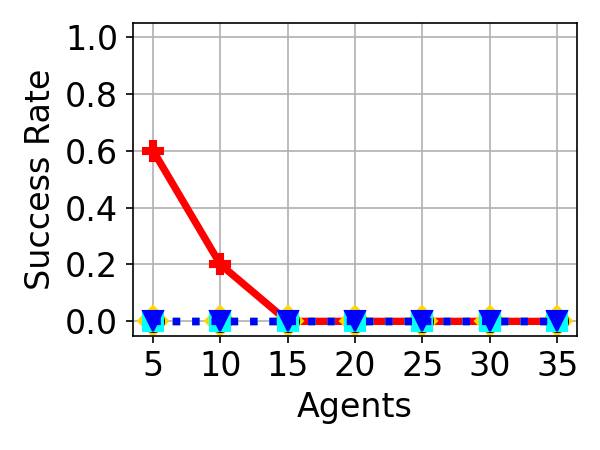}} &
    \subcaptionbox{{random-32-32-20}}[0.19\textwidth]{%
      \includegraphics[width=\linewidth]{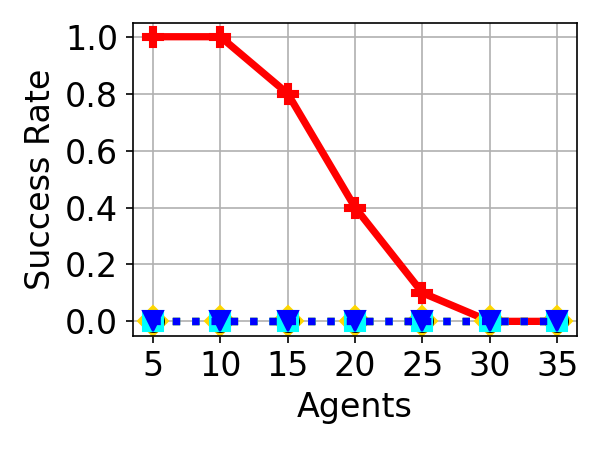}} \\
    \subcaptionbox{{random-64-64-10}}[0.19\textwidth]{%
      \includegraphics[width=\linewidth]{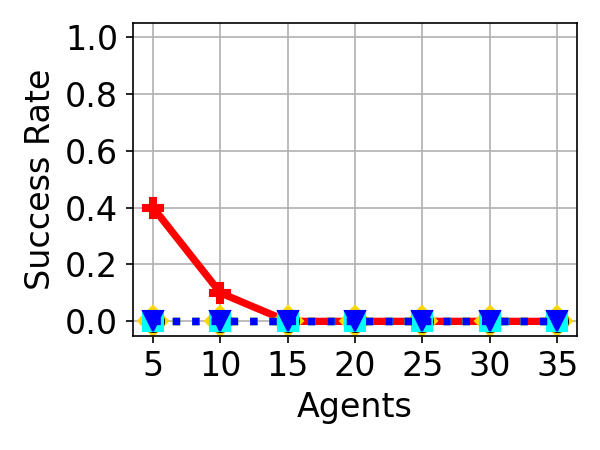}} &
    \subcaptionbox{{random-64-64-20}}[0.19\textwidth]{%
      \includegraphics[width=\linewidth]{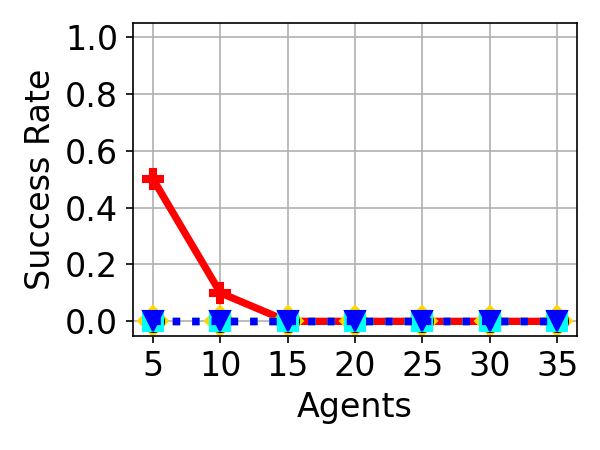}} &
    \subcaptionbox{{room-32-32-4}}[0.19\textwidth]{%
      \includegraphics[width=\linewidth]{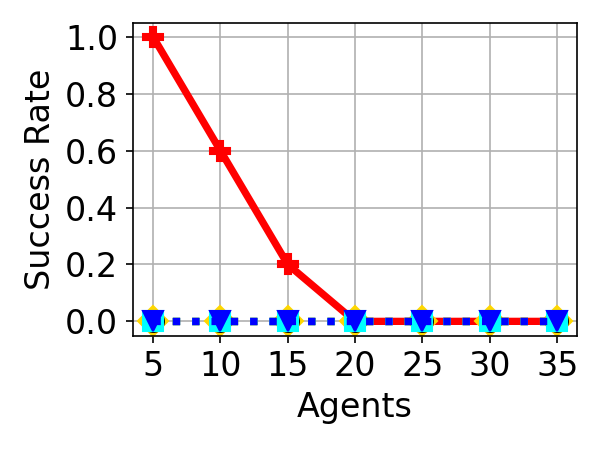}} &
    \subcaptionbox{{room-64-64-8}}[0.19\textwidth]{%
      \includegraphics[width=\linewidth]{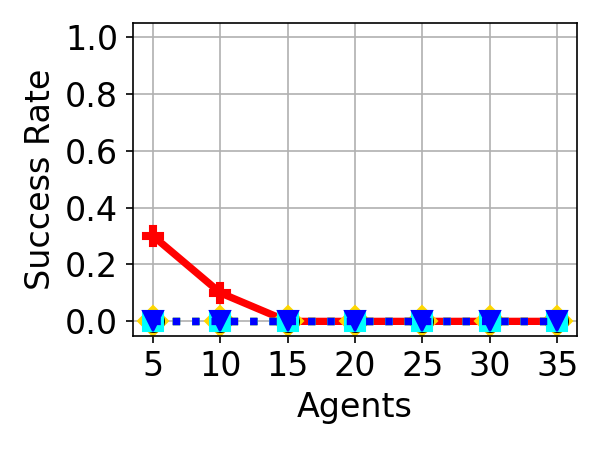}} &
    \subcaptionbox{{warehouse-10-20-10-2-1}}[0.19\textwidth]{%
      \includegraphics[width=\linewidth]{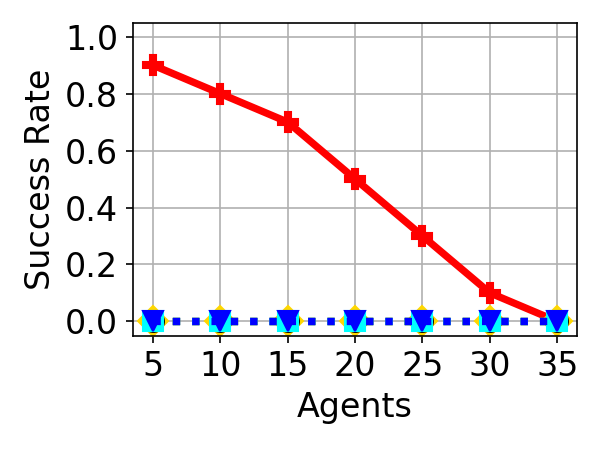}} \\
  \end{tabular}

  \caption{Success rate trends across MAPF benchmarks for 10 \emph{randomized} scenarios with \textbf{4 objectives} $(T=2\text{ mins})$.}
  \label{fig:scenario-grid-4objs_rand}
\end{figure*}

\begin{prop}\label{prop:time-d}
Runtime complexity of LCBS is linear in the number of objectives \(d\).
\end{prop}

\begin{proof}
To compute runtime complexity of LCBS, we first analyze the low level (Alg.~\ref{alg:LA*}) and the high level (Alg.~\ref{alg:LCBS}) independently, and then sum the terms. Let \(N_{\mathrm{LL}}\) be the total number of heap operations on LA\(^*\) open lists across all invocations, and \(N_{\mathrm{HL}}\) the total number of heap operations on the high-level open list. Let \(N_{\mathrm{succ}}\) be the total number of generated successors in LA\(^*\) across all calls, and \(L\) the total length of all reconstructed paths.

\textit{Low-level (Alg.~\ref{alg:LA*}).}  
Initialization (Lines 3–6) performs one push with a vector key and sets \(\mathbf{g},\mathbf{f}\), which costs \(O(d)\) for vector arithmetic and \(O(d\log|\mathcal{O}|)\) for the heap comparisons contributes \(O(d\log|\mathcal{O}|+d)\) once.  
Each iteration removes the state (defined in Line 1) with the lexicographically smallest \(\mathbf{f}\) (Line 8). A pop from a binary heap costs \(O(\log|\mathcal{O}|)\), and each lexicographic comparison on \(d\)-dimensional vectors costs \(O(d)\), so Line 8 contributes \(O(d\log|\mathcal{O}|)\) per pop, totaling \(O\!\big(d\,N_{\mathrm{LL}}\log N_{\mathrm{LL}}\big)\).  
Goal checking and path reconstruction (Lines 9–10) are independent of \(d\), costing \(O(L)\) overall.  
Successor generation (Lines 11–12) creates \(N_{\mathrm{succ}}\) candidates in total, independent of \(d\).  
Constraint checks (Lines 13–14) are independent of \(d\).
For each inserted successor (Lines 15–18), computing \(\mathbf{g}'=\mathbf{g}+\mathbf{c}_e\) and \(\mathbf{f}'=\mathbf{g}'+\mathbf{h}\) costs \(O(d)\) vector arithmetic, updating the map (Line 17) costs \(O(d)\), and the push incurs \(O(d\log|\mathcal{O}|)\). Summed over all insertions, it contributes \(O\!\big(d\,N_{\mathrm{LL}}\log N_{\mathrm{LL}}+d\,N_{\mathrm{LL}}\big)\).  
Putting these terms together, the low-level algorithm takes:
\begin{equation}
    O\!\big(d\,N_{\mathrm{LL}}\log N_{\mathrm{LL}} + d\,N_{\mathrm{LL}}\big) \;+\; O(N_{\mathrm{succ}} + L).
    \label{eqn:low-level}
\end{equation}

\textit{High-level (Alg.~\ref{alg:LCBS}).}  
Root construction (Lines 1–7) calls LA\(^*\) once per agent, already included in the low-level accounting. Computing \(\mathcal{N}_0.\mathbf{C}=\sum_i \mathbf{c}(\pi_i)\) (Line 6) costs \(O(d)\) per agent and costs \(O(d|A|)\). Inserting the root (Line 8) in the heap costs \(O(d\log|\mathcal{O}_{\mathrm{HL}}|)\).  
Each iteration removes the node with the lexicographically smallest joint cost vector (Line 10), which costs \(O(d\log|\mathcal{O}_{\mathrm{HL}}|)\) per pop, totaling \(O\!\big(d\,N_{\mathrm{HL}}\log N_{\mathrm{HL}}\big)\).  
Conflict detection (Line 11) and constraint generation (Line 16) are independent of \(d\).  
Replanning the constrained agent (Line 19) invokes LA\(^*\), already accounted for in the low-level analysis.
For each feasible child (Lines 21–24), updating \(\mathcal{N}_a.\mathbf{C}\) after replacing one agent’s path, costs \(O(d)\), and inserting the child into the heap costs \(O(d\log|\mathcal{O}_{\mathrm{HL}}|)\). Summed over all inserted children, it amounts to \(O\!\big(d\,N_{\mathrm{HL}}\log N_{\mathrm{HL}}+d\,N_{\mathrm{HL}}\big)\). Thus, the high-level algorithm takes 
\begin{equation}
    O\!\big(d\,N_{\mathrm{HL}}\log N_{\mathrm{HL}} + d\,N_{\mathrm{HL}}\big) \;+\; O(1 + |A| + C),
     \label{eqn:high-level}
\end{equation}
where \(C\) denotes the contribution of conflict detection and constraint generation, independent of \(d\).

From Equations~\ref{eqn:low-level} and~\ref{eqn:high-level}, we get
\begin{equation*}
\begin{split}
O\!\big(d\,(N_{\mathrm{LL}}\log N_{\mathrm{LL}} + N_{\mathrm{LL}}
+ N_{\mathrm{HL}}\log N_{\mathrm{HL}} + N_{\mathrm{HL}})\big) \\
{}+ O(N_{\mathrm{succ}} + L + |A| + C + 1)
\end{split}
\end{equation*}

\begin{equation*}
\begin{split}
\le\; O\!\Big(d\big[(N_{\mathrm{LL}}+N_{\mathrm{HL}})\log(N_{\mathrm{LL}}+N_{\mathrm{HL}}) + (N_{\mathrm{LL}}+N_{\mathrm{HL}})\big]\Big)
\\
{}+ O(N_{\mathrm{succ}} + L + |A| + C + 1),
\end{split}
\end{equation*}

\begin{equation*}
\begin{split}
=\; O\!\big(d\,N\log N\big) \;+\; O(N_{\mathrm{succ}} + L + |A| + C + 1),
\\
{} \text{where, } N:=N_{\mathrm{LL}}+N_{\mathrm{HL}},
\end{split}
\end{equation*}

\begin{equation*}
\begin{split}
= \; O(d)\,F(N) \;+\; O(N_{\mathrm{succ}} + L + |A| + C + 1),
\\
{} \text{where, } F(N):=O\!\big(N\log N\big),
\end{split}
\end{equation*}
showing linear dependence on $d$. Therefore, runtime complexity of LCBS is linear in the number of objectives \(d\).
\end{proof}

\begin{table*}[t]
\renewcommand{\arraystretch}{1.1}
\centering
\setlength{\tabcolsep}{3pt}
\footnotesize

\begin{tabularx}{\textwidth}{|>{\centering\arraybackslash}m{0.18\textwidth}||*{6}{>{\centering\arraybackslash}X|}}
\hline
\multirow{2}{*}{\textbf{Benchmark}} & \multicolumn{6}{c|}{\textbf{Total Cost Vectors from Representative Scenarios (5 agents, 3 objectives)}} \\ \cline{2-7}
& \textbf{LCBS (ours)} & \textbf{BB-k (1)} & \textbf{BB-k (5)} & \textbf{BB-k (10)} & \textbf{BB-pex} & \textbf{BB-$\epsilon$} \\
\specialrule{1.1pt}{0pt}{0pt}
\text{empty-32-32}      & \(\{137, 268, 182\}\) & \(\{137, 268, 182\}\) & \(\{137, 268, 182\}\) & \(\{137, 268, 182\}\) & \(\{137, 268, 182\}\) & \(\{137, 268, 182\}\) \\ \hline
\text{empty-48-48}      & \(\{148, 261, 183\}\) & \(\{148, 261, 183\}\) & \(\{148, 261, 183\}\) & \(\{148, 261, 183\}\) & \(\{148, 261, 183\}\) & \(\{148, 261, 183\}\) \\ \hline
\text{maze-32-32-2}     & \(\{86, 188, 237\}\)  & \(\{86, 188, 237\}\)  & \(\{86, 188, 237\}\)  & \(\{86, 188, 237\}\)  & \(\{86, 188, 237\}\)  & \(\{86, 188, 237\}\)  \\ \hline
\text{maze-32-32-4}     & \(\{106, 188, 217\}\) & \(\{106, 188, 217\}\) & \(\{106, 188, 217\}\) & \(\{106, 188, 217\}\) & \(\{106, 188, 217\}\) & \(\{106, 188, 217\}\) \\ \hline
\text{random-32-32-20}  & \(\{221, 239, 229\}\) & \(\{221, 239, 229\}\) & \(\{221, 239, 229\}\) & \(\{221, 239, 229\}\) & \(\{221, 239, 229\}\) & \(\{221, 239, 229\}\) \\ \hline
\text{random-64-64-10}  & \(\{201, 267, 298\}\) & \(\{201, 267, 298\}\) & \(\{201, 267, 298\}\) & \(\{201, 267, 298\}\) & \(\{201, 267, 298\}\) & \(\{201, 267, 298\}\) \\ \hline
\text{random-64-64-20}  & \(\{224, 289, 276\}\) & \(\{224, 289, 276\}\) & \(\{224, 289, 276\}\) & \(\{224, 289, 276\}\) & \(\{224, 289, 276\}\) & \(\{224, 289, 276\}\) \\ \hline
\text{room-32-32-4}     & \(\{149, 209, 157\}\) & \(\{149, 209, 157\}\) & \(\{149, 209, 157\}\) & \(\{149, 209, 157\}\) & \(\{149, 209, 157\}\) & \(\{149, 209, 157\}\) \\ \hline
\text{room-64-64-8}     & \(\{132, 98, 226\}\)  & \(\{132, 98, 226\}\)  & \(\{132, 98, 226\}\)  & \(\{132, 98, 226\}\)  & \(\{132, 98, 226\}\)  & \(\{132, 98, 226\}\)  \\ \hline
\text{warehouse-10-20-10-2-1}     & \(\{104, 119, 95\}\)  & \(\{104, 119, 95\}\)  & \(\{104, 119, 95\}\)  & \(\{104, 119, 95\}\)  & \(\{104, 119, 95\}\)  & \(\{104, 119, 95\}\)  \\ \hline
\end{tabularx}

\caption{Empirical verification that LCBS computes same total-cost vectors as Pareto-optimal baselines (BB:= BB-MO-CBS).}
\label{tab:optimality-comparison}
\end{table*}

\section{Experiments}
We evaluate LCBS against five baselines across ten MAPF benchmarks~\cite{stern2019mapf} with standard and randomized scenarios. We use the baseline implementations from the codebase of~\citet{wang2024efficient}. All algorithms were implemented in C++ and tested on a macOS machine with Apple M3 Pro chip and $18$ GB RAM.

\paragraph{Baselines} 
We compare LCBS against four variants proposed by~\citet{wang2024efficient}, all of which use the same high-level planner as LCBS but differ in low-level search. Each method reports a single plan aligned with the lexicographic order. If a method returns multiple candidates, we choose the plan that is lexicographically optimal using Equation \ref{eq:lex-optimality}.

\begin{itemize}[noitemsep,nolistsep,leftmargin=*] \item \emph{BB-MO-CBS-$k$ $(k=1)$}: Returns a single non-dominated solution using adaptive $\varepsilon$; provides a fast Pareto baseline with minimal computation. \item \emph{BB-MO-CBS-$k$ $(k=5)$}: Returns five solutions to evaluate whether modest Pareto coverage improves success. \item \emph{BB-MO-CBS-$k$ $(k=10)$}: Returns ten solutions to test the trade-off between Pareto exploration and runtime. \item \emph{BB-MO-CBS-pex $(\varepsilon=0.03)$}: Reuses joint paths and merges duplicate conflicts to reduce computational load. \end{itemize}
In addition, we compare with \emph{BB-MO-CBS-$\varepsilon$} $(\varepsilon=0.03)$~\cite{ren2023binary}, which differs in high-level and low-level planning. It prunes $\varepsilon$-dominated paths for speed-up.

\paragraph{Parameter choices} Similar to prior work~\cite{ren2023binary,wang2024efficient}, we use $\varepsilon=0.03$ for BB-MO-CBS-$\varepsilon$ and BB-MO-CBS-pex. This value controls the granularity of $\varepsilon$-dominance and is the standard choice in those baselines. For BB-MO-CBS-$k$ we report $k\in\{1,5,10\}$ to cover low, modest, and broader set coverage under the same time limits, matching the ranges evaluated by~\citet{wang2024efficient}.

We compare the algorithms in terms of: (i) success rates as the number of agents increases, (ii) performance trends across increasing numbers of objectives, and (iii) optimality.

\section{Results and Discussion}

\paragraph{Success rates}
We evaluate each method's ability to find a valid joint plan within a time limit as the number of agents increase. For fair comparison with Pareto-based approaches and following existing literature, a scenario is considered a \emph{success} if at least one valid solution is found within \emph{two} minutes~\cite{wang2024efficient}. Each scenario defines start and goal locations for all agents. Evaluations are conducted on \emph{ten} benchmark maps, each with $25$ standard scenarios~\cite{stern2019mapf} and $10$ randomized ones. Success rate denotes the fraction of scenarios that were a success.

Results in Fig.~\ref{fig:scenario-grid-3objs} show that LCBS achieves higher success rates across all benchmarks for standard scenarios with three objectives. In some cases, LCBS is able to solve instances with up to $35$ agents, whereas baselines fail beyond $25$. With four objectives, as seen in Fig.~\ref{fig:scenario-grid-4objs}, LCBS is the \emph{only} approach that produces any success; the baselines fail due to the growing overhead of approximating the Pareto frontier. In some cases, LCBS solves all scenarios where others solve none, highlighting its utility in more demanding settings. We observe similar trends for randomized scenarios in Fig.~\ref{fig:scenario-grid-rand-3objs} and ~\ref{fig:scenario-grid-4objs_rand}, across both three and four objectives, showing robust performance.

\begin{figure*}[!h]
  \centering

  \begin{tabular}{@{}m{0.045\textwidth}*{6}{m{0.153\textwidth}}@{}}
    \rotatebox{90}{\small (a) \emph{empty-48-48}} &
    \methodpanel{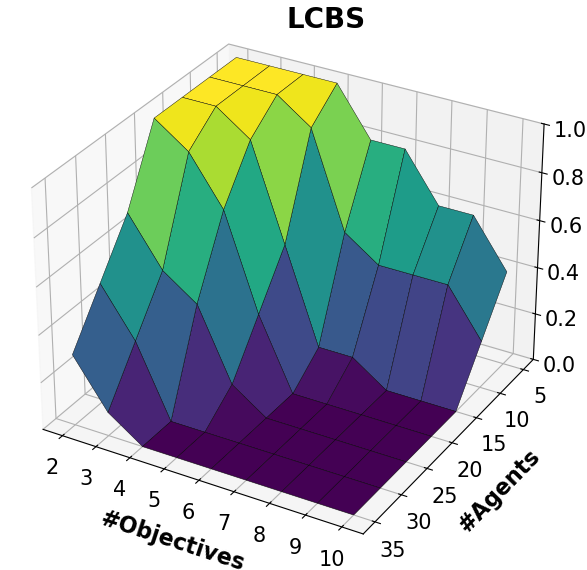}{LCBS (ours)} &
    \methodpanel{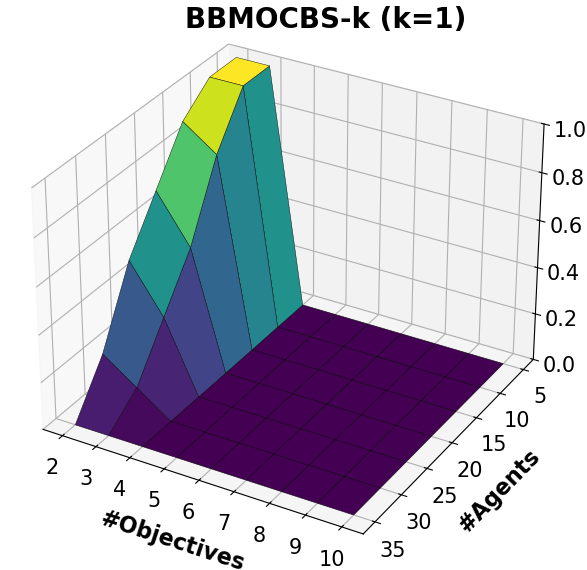}{BB-MO-CBS-k (k=1)} &
    \methodpanel{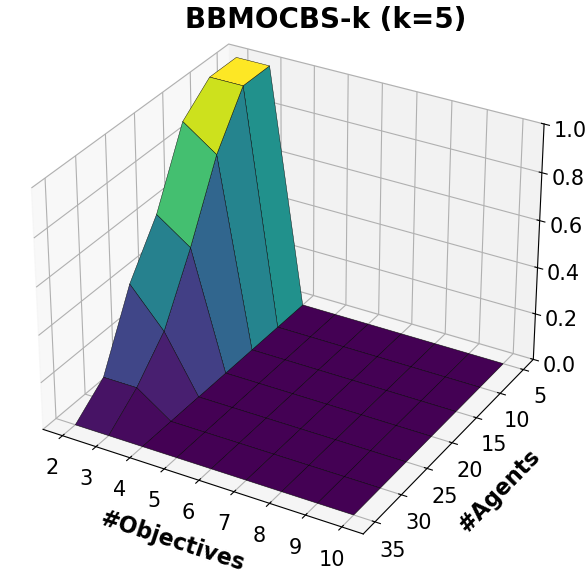}{BB-MO-CBS-k (k=5)} &
    \methodpanel{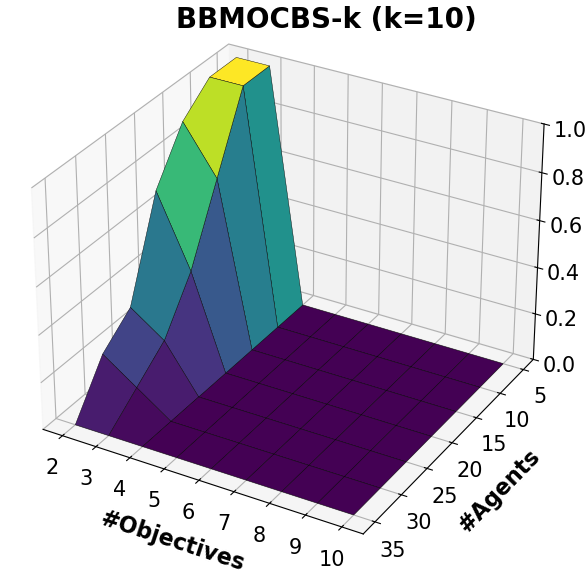}{BB-MO-CBS-k (k=10)} &
    \methodpanel{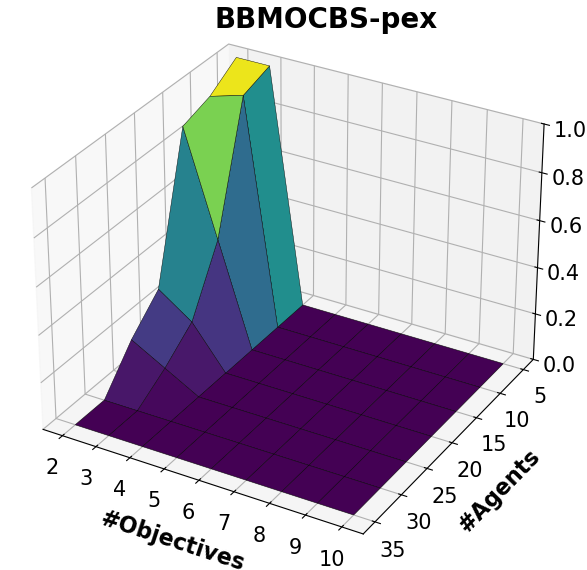}{BB-MO-CBS-pex} &
    \methodpanel{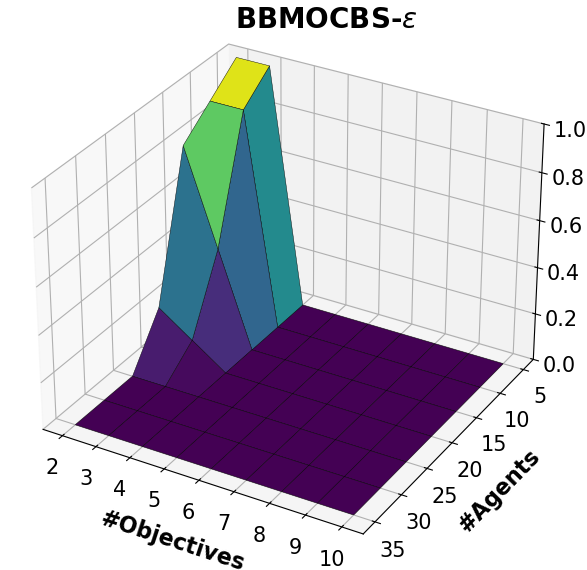}{BB-MO-CBS-$\epsilon$ ($\epsilon{=}0.03$)} \\
  \end{tabular}

  \vspace{6pt}

  \begin{tabular}{@{}m{0.045\textwidth}*{6}{m{0.153\textwidth}}@{}}
    \rotatebox{90}{\small (b) \emph{random-64-64-10}} &
    \methodpanel{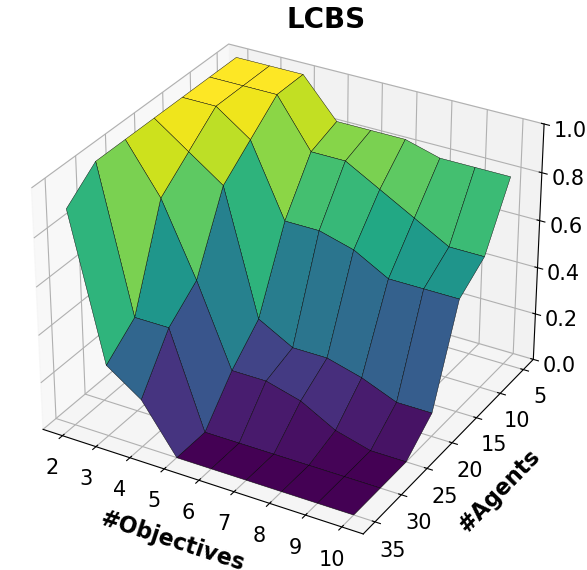}{LCBS (ours)} &
    \methodpanel{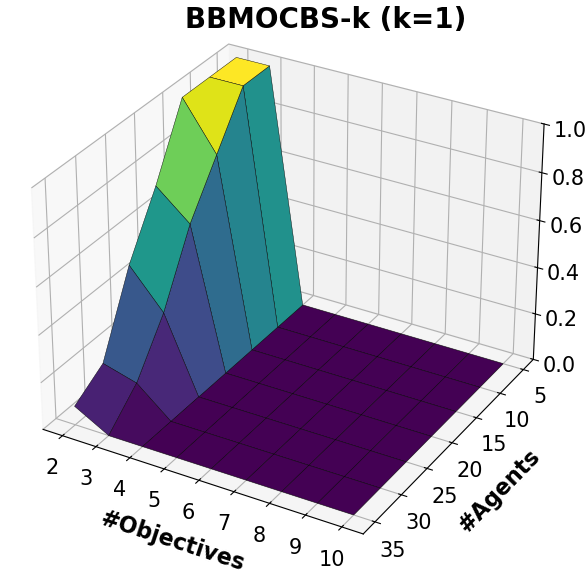}{BB-MO-CBS-k (k=1)} &
    \methodpanel{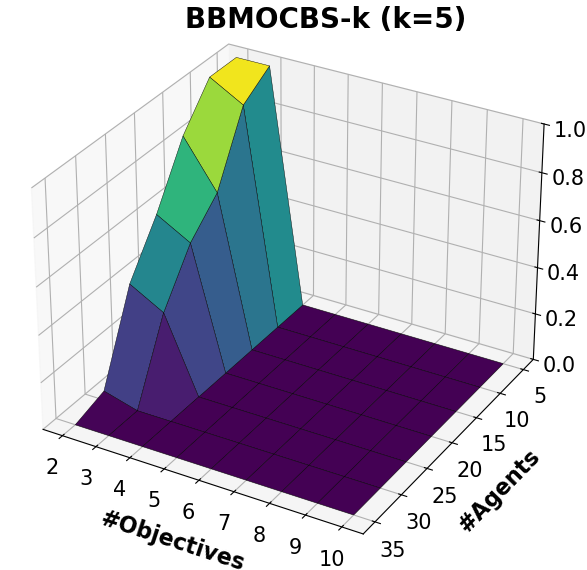}{BB-MO-CBS-k (k=5)} &
    \methodpanel{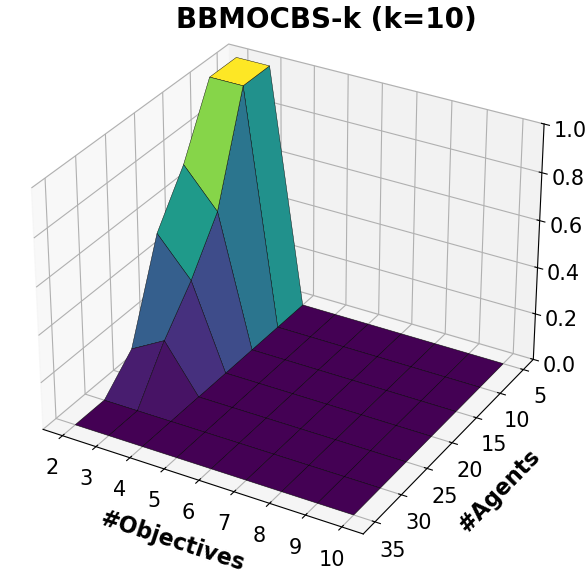}{BB-MO-CBS-k (k=10)} &
    \methodpanel{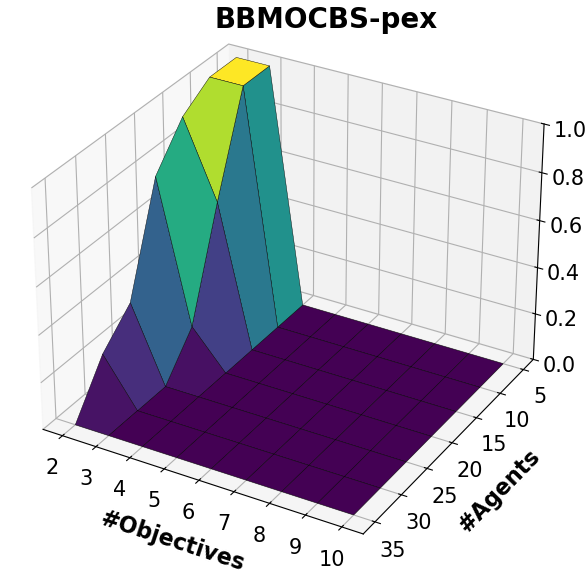}{BB-MO-CBS-pex} &
    \methodpanel{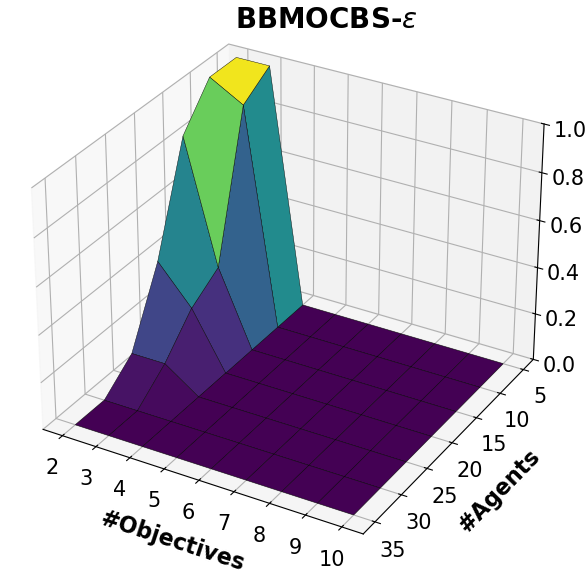}{BB-MO-CBS-$\epsilon$ ($\epsilon{=}0.03$)} \\
  \end{tabular}

  \vspace{6pt}

  \begin{tabular}{@{}m{0.045\textwidth}*{6}{m{0.153\textwidth}}@{}}
    \rotatebox{90}{\small (c) \emph{random-64-64-20}} &
    \methodpanel{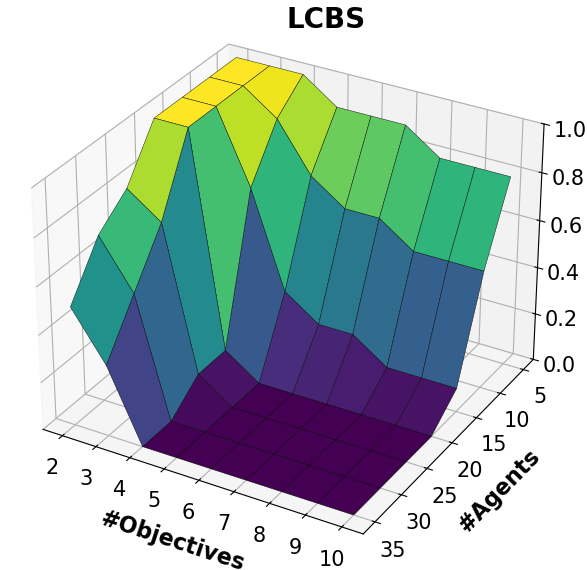}{LCBS (ours)} &
    \methodpanel{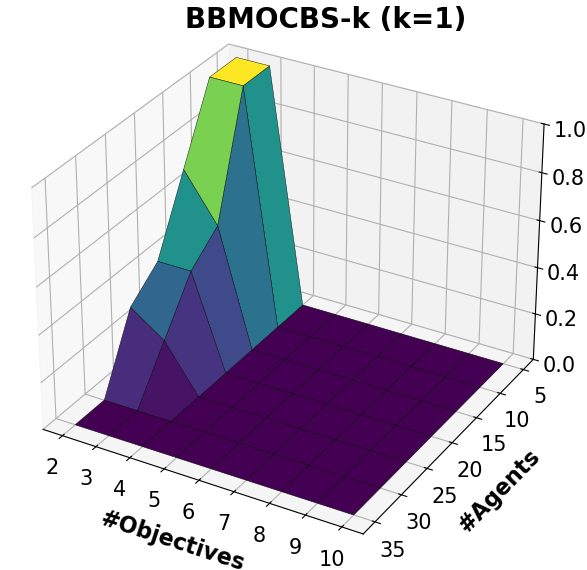}{BB-MO-CBS-k (k=1)} &
    \methodpanel{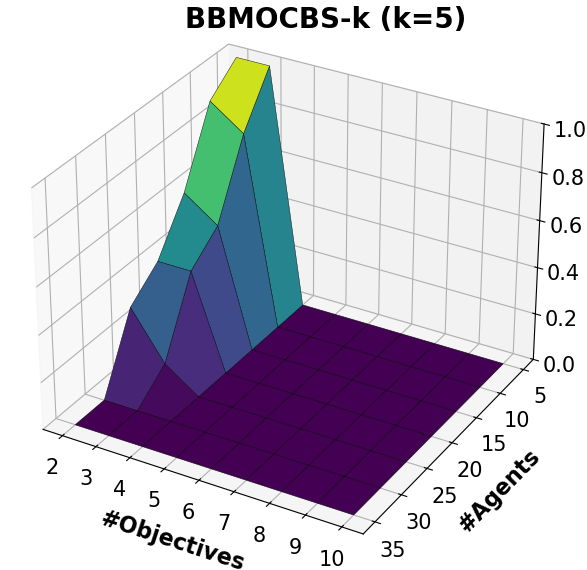}{BB-MO-CBS-k (k=5)} &
    \methodpanel{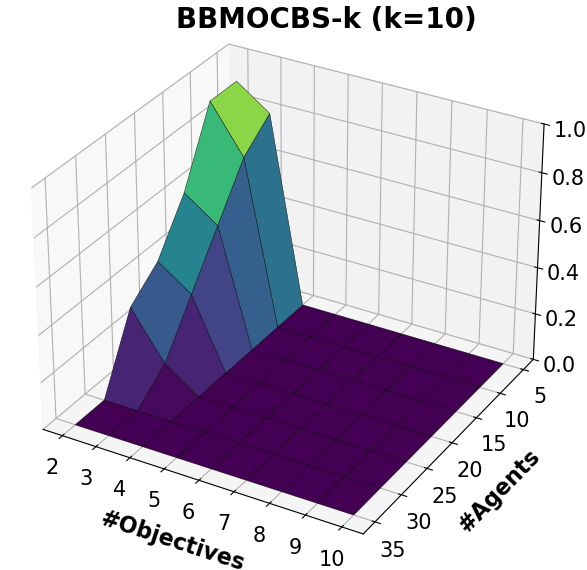}{BB-MO-CBS-k (k=10)} &
    \methodpanel{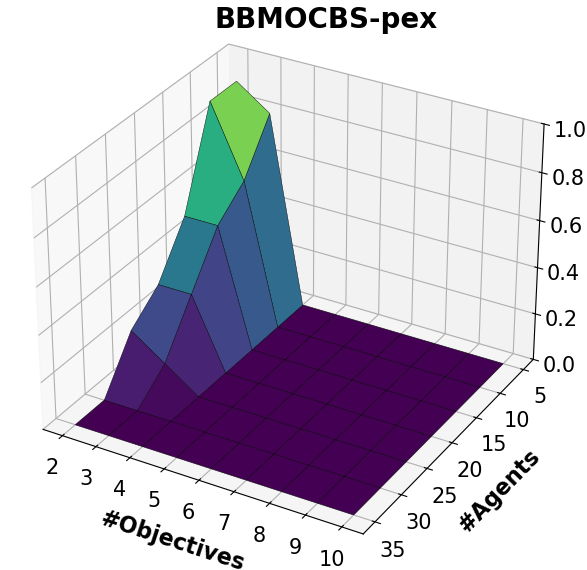}{BB-MO-CBS-pex} &
    \methodpanel{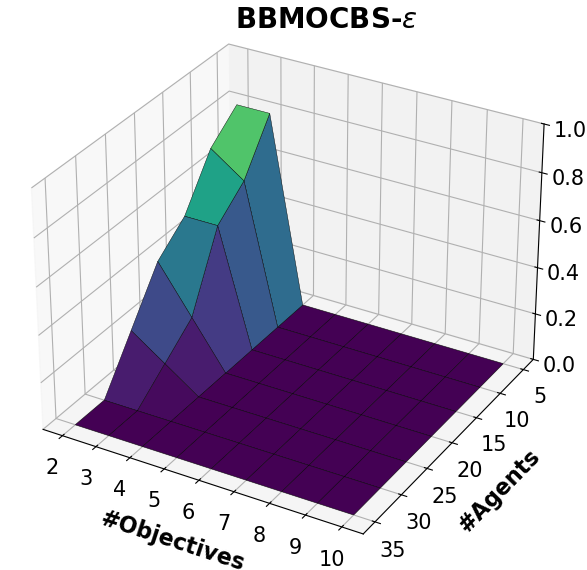}{BB-MO-CBS-$\epsilon$ ($\epsilon{=}0.03$)} \\
  \end{tabular}

  \vspace{6pt}

  \begin{tabular}{@{}m{0.045\textwidth}*{6}{m{0.153\textwidth}}@{}}
    \rotatebox{90}{\small (d) \emph{room-32-32-4}} &
    \methodpanel{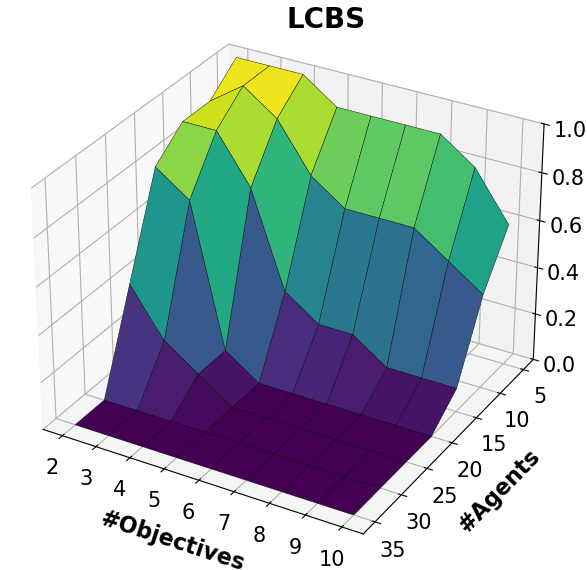}{LCBS (ours)} &
    \methodpanel{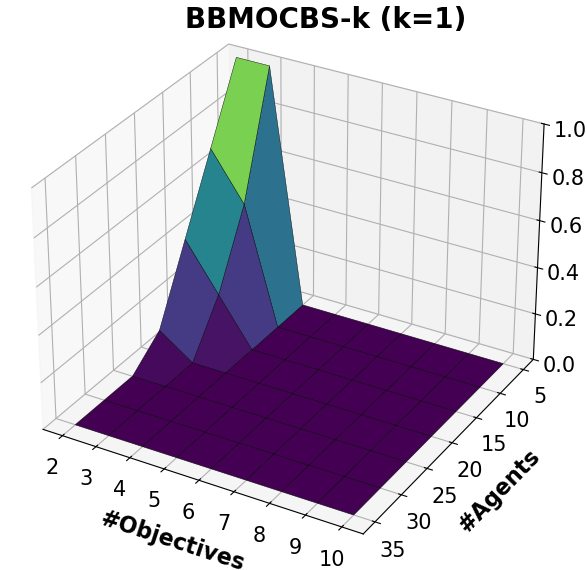}{BB-MO-CBS-k (k=1)} &
    \methodpanel{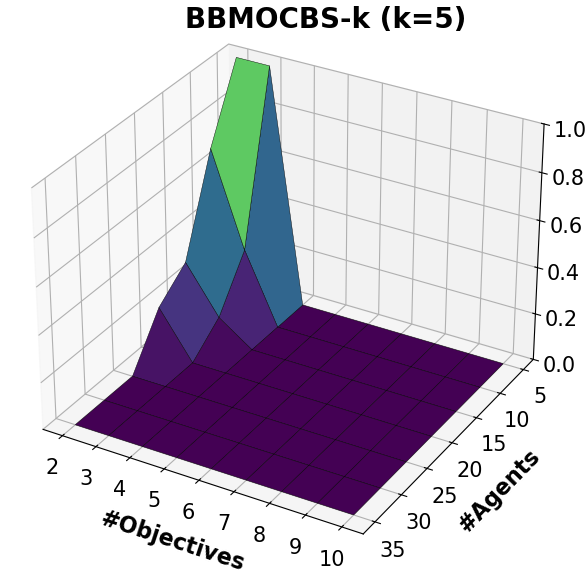}{BB-MO-CBS-k (k=5)} &
    \methodpanel{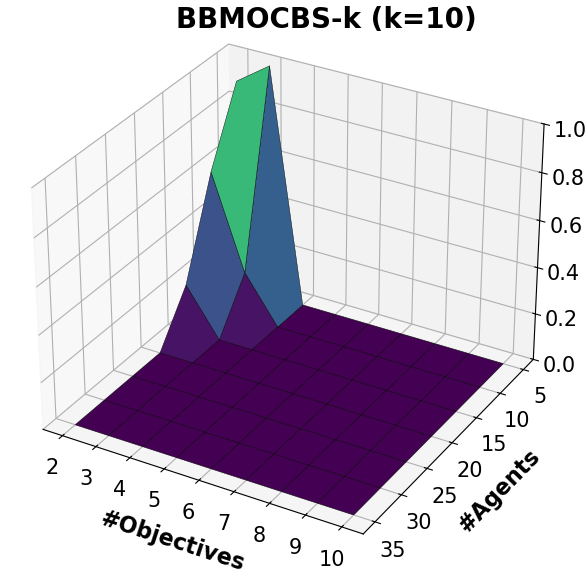}{BB-MO-CBS-k (k=10)} &
    \methodpanel{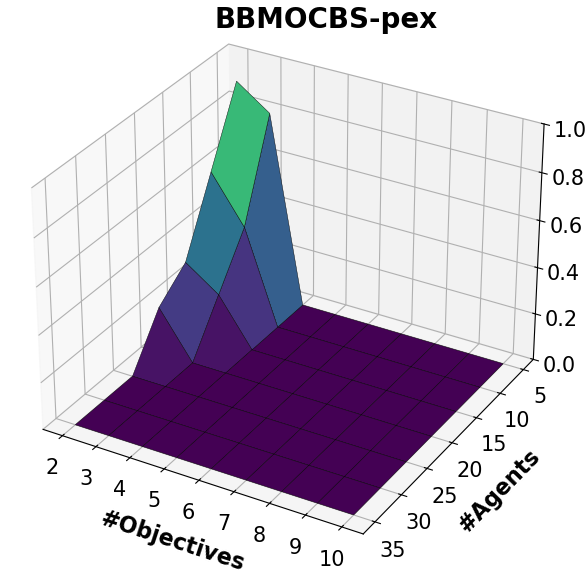}{BB-MO-CBS-pex} &
    \methodpanel{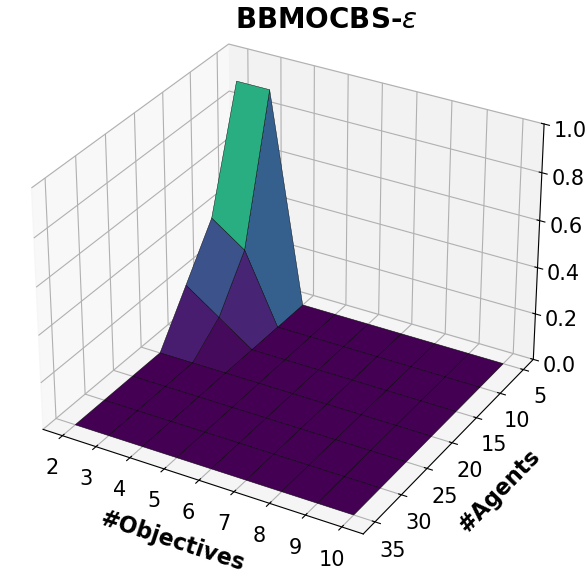}{BB-MO-CBS-$\epsilon$ ($\epsilon{=}0.03$)} \\
  \end{tabular}
  \caption{Success-contour grids: each row is a benchmark (left, vertical label) and columns are methods (subcaptions).}
  \label{fig:success-rate-contours-grid-1}
\end{figure*}

\paragraph{Scaling with number of objectives} 
We test scalability of all approaches up to $10$ objectives on all benchmarks. Each benchmark is tested on $25$ standard scenarios~\cite{stern2019mapf} with a $5$-minute time limit to accommodate added complexity. Figure~\ref{fig:success-rate-contours-grid-1} shows results on a subset of the benchmark environments. LCBS is the \emph{only} method that produces any success beyond $3$ objectives, scaling up to $10$ objectives. The baselines fail beyond three objectives, primarily due to the overhead of approximating the Pareto frontier. 
These results show that LCBS scales effectively with the number of objectives, demonstrating its practical utility for real-world applications with numerous objectives.

\paragraph{Solution optimality}
To evaluate whether LCBS returns lexicographically optimal solutions, we compare its outputs against methods established to compute the Pareto frontier~\cite{wang2024efficient,ren2023binary}. For Pareto-based methods, we select the solution that aligns with the same lexicographic priority used by LCBS. Table~\ref{tab:optimality-comparison} shows that LCBS produces the same total cost vectors for \emph{representative scenarios} as the Pareto-optimal baselines. Representative scenarios are the ones where all approaches are successful. These results confirm that LCBS returns the same optimal solution without computing the Pareto front.

\section{Related Works}

\paragraph{Multi-objective path planning} addresses settings with multiple, often conflicting, objectives. Algorithms such as NAMOA$^*$~\cite{mandow2008multiobjective} and A$^*$pex~\cite{zhang2022pex} extend A$^*$ to compute the entire Pareto frontier, but scale poorly with the number of objectives. Approximate variants trade off optimality for lower runtime, but do not guarantee Pareto-optimal solutions~\cite{ulloa2020simple,goldin2021approximate}. In contrast, we avoid computing the Pareto front by directly optimizing for user preferences, enabling scalable planning.

\paragraph{MO-MAPF algorithms} Existing MO-MAPF algorithms compute 
Pareto-optimal paths per agent via a low-level search and use a high-level planner to resolve conflicts by imposing constraints. MO-CBS~\cite{ren2021multi} constructs the full Pareto front by enumerating all non-dominated paths and resolving conflicts using a constraint tree~\cite{sharon2015conflict}. BB-MO-CBS~\cite{ren2023binary} improves conflict resolution while retaining the same low-level frontier construction. Variants such as BB-MO-CBS-$\varepsilon$~\cite{ren2023binary} and BB-MO-CBS-pex~\cite{wang2024efficient} approximate the frontier via $\varepsilon$-dominance, with the latter reducing overhead by merging redundant paths and conflicts. BB-MO-CBS-$k$~\cite{wang2024efficient} returns exactly $k$ Pareto solutions using an adaptive $\varepsilon$. Scalarization-based methods~\cite{ho2023preference}  combine multiple objectives into a single cost before applying standard CBS. 

\paragraph{Limitations and Scope}
Lexicographic framework reflects a modeling choice, not a limitation of LCBS. It provides an intuitive and scalable way to handle trade-offs in scenarios with objective preferences, optimized sequentially from highest to lowest priority~\cite{wray2015multi,xue2025multiple}. This work focuses on MO-MAPF instances where a preference order is specified and should guide planning. It does not aim to cover all MO-MAPF formulations and is not a replacement for Pareto-front approaches used when preferences are unknown.

\section{Conclusion}
We introduce a lexicographic MO-MAPF formulation and present lexicographic CBS (LCBS) algorithm that can optimally plan for lexicographic preference over objectives. By integrating priority-aware low-level search using LA$^*$ into the CBS framework, LCBS avoids computing Pareto frontiers or scalarization, enabling efficient planning guided by preference ordering. Our extensive evaluation shows that LCBS consistently outperforms baselines with higher success rates across benchmarks, and scaling to larger agent teams and higher number of objectives. 
\section{Acknowledgments}
This work was supported in part by ONR award N00014-23-1-2171 and NSF award 2416459.

\bibliography{aaai2026}

\end{document}


\maketitle
\appendix

\section{Scaling with Number of Objectives}
We evaluate scalability up to ten objectives on additional MAPF benchmarks, each with 25 standard scenarios and $5$ mins time limit. Figures~1 and 2 show that LCBS is able to produce high success rates up to ten objectives where as all other approaches fail beyond three. This highlights the dominance of LCBS over existing methods, especially in scenarios where planning involves numerous objectives.
\begin{center}
\vspace{25pt}
\begin{minipage}{\textwidth}
    \begin{minipage}[t]{0.02\textwidth}
        \centering
        \rotatebox{90}{\small(a) \emph{empty-32-32}}
    \end{minipage}%
    \hfill
    \begin{minipage}[t]{0.96\textwidth}
        \centering
        \begin{minipage}[t]{0.162\textwidth}
            \includegraphics[width=\linewidth,trim={1.1cm 0 0 1.1cm},clip]{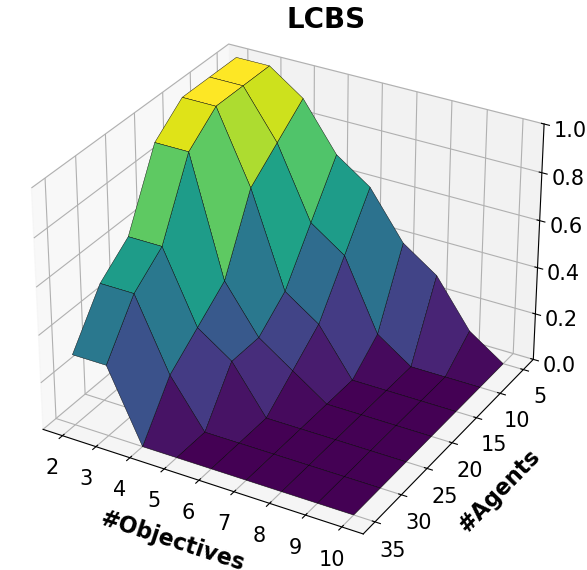} \\
            \centering \scriptsize LCBS (ours)
        \end{minipage}
        \hfill
        \begin{minipage}[t]{0.162\textwidth}
            \includegraphics[width=\linewidth,trim={1.1cm 0 0 1.1cm},clip]{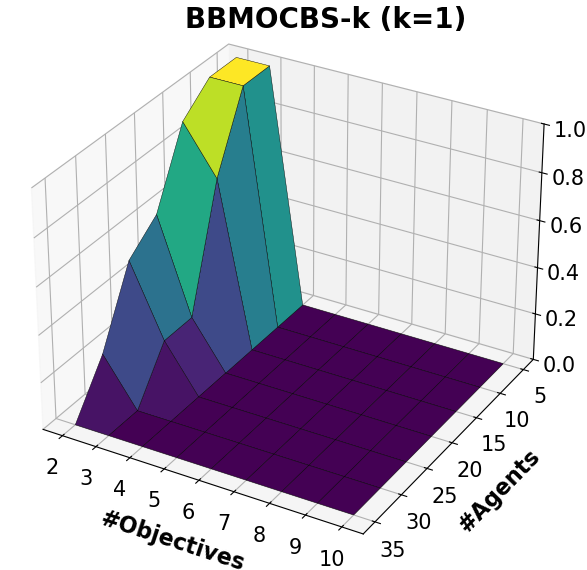} \\
            \centering \scriptsize BB-MO-CBS-k (k=1)
        \end{minipage}
        \hfill
        \begin{minipage}[t]{0.162\textwidth}
            \includegraphics[width=\linewidth,trim={1.1cm 0 0 1.1cm},clip]{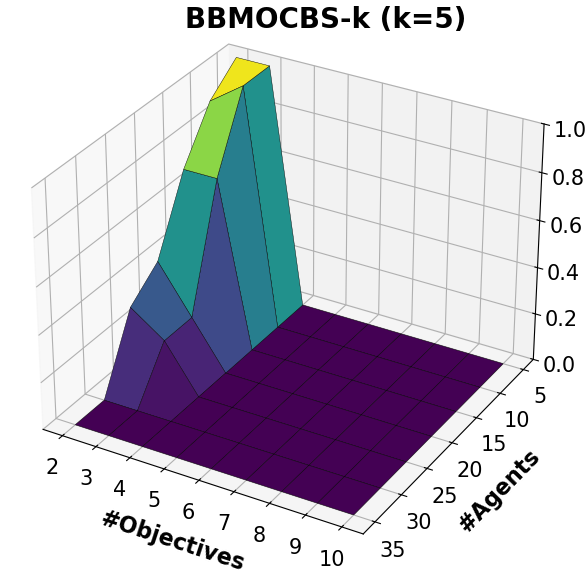} \\
            \centering \scriptsize BB-MO-CBS-k (k=5)
        \end{minipage}
        \hfill
        \begin{minipage}[t]{0.162\textwidth}
            \includegraphics[width=\linewidth,trim={1.1cm 0 0 1.1cm},clip]{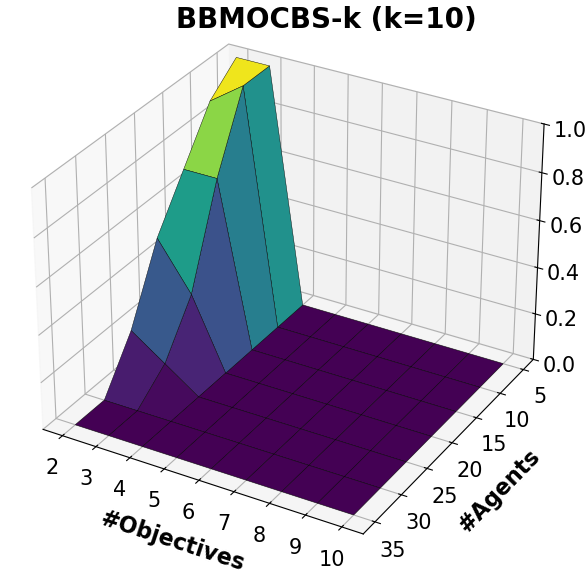} \\
            \centering \scriptsize BB-MO-CBS-k (k=10)
        \end{minipage}
        \hfill
        \begin{minipage}[t]{0.162\textwidth}
            \includegraphics[width=\linewidth,trim={1.1cm 0 0 1.1cm},clip]{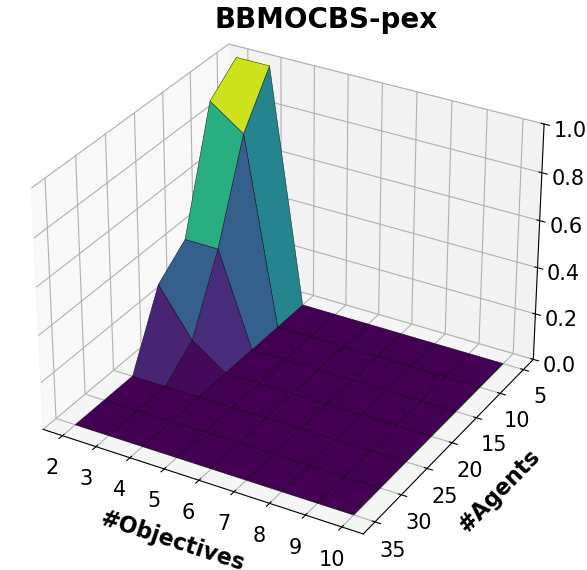} \\
            \centering \scriptsize BB-MO-CBS-pex
        \end{minipage}
        \hfill
        \begin{minipage}[t]{0.162\textwidth}
            \includegraphics[width=\linewidth,trim={1.1cm 0 0 1.1cm},clip]{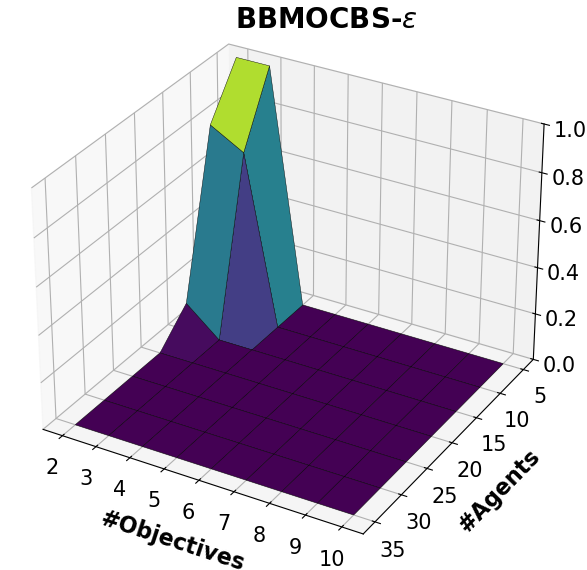} \\
            \centering \scriptsize BB-MO-CBS-$\epsilon$
        \end{minipage}
    \end{minipage}
\end{minipage}
\end{center}

\begin{center}
\begin{minipage}[t]{\textwidth}
     \begin{minipage}[t]{0.02\textwidth}
        \centering
        \rotatebox{90}{\small(b) \emph{random-64-64-10}}
    \end{minipage}%
    \hfill
    \begin{minipage}[t]{0.96\textwidth}
        \centering
        \begin{minipage}[t]{0.1625\textwidth}
            \includegraphics[width=\linewidth,trim={1.1cm 0 0 1.1cm},clip]{figures/random-64-64-10/LCBS.png} \\
            \centering \scriptsize LCBS (ours)
        \end{minipage}
        \hfill
        \begin{minipage}[t]{0.1625\textwidth}
            \includegraphics[width=\linewidth,trim={1.1cm 0 0 1.1cm},clip]{figures/random-64-64-10/BBMOCBS-k_1.png} \\
            \centering \scriptsize BB-MO-CBS-k (k=1)
        \end{minipage}
        \hfill
        \begin{minipage}[t]{0.1625\textwidth}
            \includegraphics[width=\linewidth,trim={1.1cm 0 0 1.1cm},clip]{figures/random-64-64-10/BBMOCBS-k_5.png} \\
            \centering \scriptsize BB-MO-CBS-k (k=5)
        \end{minipage}
        \hfill
        \begin{minipage}[t]{0.1625\textwidth}
            \includegraphics[width=\linewidth,trim={1.1cm 0 0 1.1cm},clip]{figures/random-64-64-10/BBMOCBS-k_10.png} \\
            \centering \scriptsize BB-MO-CBS-k (k=10)
        \end{minipage}
        \hfill
        \begin{minipage}[t]{0.1625\textwidth}
            \includegraphics[width=\linewidth,trim={1.1cm 0 0 1.1cm},clip]{figures/random-64-64-10/BBMOCBS-pex.png} \\
            \centering \scriptsize BB-MO-CBS-pex
        \end{minipage}
        \hfill
        \begin{minipage}[t]{0.1625\textwidth}
            \includegraphics[width=\linewidth,trim={1.1cm 0 0 1.1cm},clip]{figures/random-64-64-10/BBMOCBS-eps.png}\\
            \centering \scriptsize BB-MO-CBS-$\epsilon$
        \end{minipage}
    \end{minipage}
\end{minipage}
\end{center}

\begin{center}
\begin{minipage}[t]{\textwidth}
     \begin{minipage}[t]{0.02\textwidth}
        \centering
        \rotatebox{90}{\small(c) \emph{random-64-64-20}}
    \end{minipage}%
    \hfill
    \begin{minipage}[t]{0.96\textwidth}
        \centering
        \begin{minipage}[t]{0.1625\textwidth}
            \includegraphics[width=\linewidth,trim={1.1cm 0 0 1.1cm},clip]{figures/random-64-64-20/LCBS.png} \\
            \centering \scriptsize LCBS (ours)
        \end{minipage}
        \hfill
        \begin{minipage}[t]{0.1625\textwidth}
            \includegraphics[width=\linewidth,trim={1.1cm 0 0 1.1cm},clip]{figures/random-64-64-20/BBMOCBS-k_1.png} \\
            \centering \scriptsize BB-MO-CBS-k (k=1)
        \end{minipage}
        \hfill
        \begin{minipage}[t]{0.1625\textwidth}
            \includegraphics[width=\linewidth,trim={1.1cm 0 0 1.1cm},clip]{figures/random-64-64-20/BBMOCBS-k_5.png} \\
            \centering \scriptsize BB-MO-CBS-k (k=5)
        \end{minipage}
        \hfill
        \begin{minipage}[t]{0.1625\textwidth}
            \includegraphics[width=\linewidth,trim={1.1cm 0 0 1.1cm},clip]{figures/random-64-64-20/BBMOCBS-k_10.png} \\
            \centering \scriptsize BB-MO-CBS-k (k=10)
        \end{minipage}
        \hfill
        \begin{minipage}[t]{0.1625\textwidth}
            \includegraphics[width=\linewidth,trim={1.1cm 0 0 1.1cm},clip]{figures/random-64-64-20/BBMOCBS-pex.png} \\
            \centering \scriptsize BB-MO-CBS-pex
        \end{minipage}
        \hfill
        \begin{minipage}[t]{0.1625\textwidth}
            \includegraphics[width=\linewidth,trim={1.1cm 0 0 1.1cm},clip]{figures/random-64-64-20/BBMOCBS-eps.png}\\
            \centering \scriptsize BB-MO-CBS-$\epsilon$
        \end{minipage}
    \end{minipage}  
    \centering
    
    {\vspace{10pt}\normalsize Figure 1: Success contours for MAPF benchmarks with 25 standard scenarios each ($T=5$ min)}
\end{minipage}
\end{center}

\newpage
\section{Solution Optimality}
We evaluate solution quality on scenarios where all approaches succeed, varying the number of agents and objectives to show that optimality is a property of LCBS and not a consequence of specific system parameters. Tables~1--5 show that LCBS returns the same solutions as Pareto-based methods. Note that all cost vectors are infact identical across approaches and have been verified. We restrict this comparison up to three objectives, as baselines fail to scale beyond that.

\newpage
\begin{center}
\begin{minipage}[t]{\textwidth}
     \begin{minipage}[t]{0.02\textwidth}
        \centering
        \rotatebox{90}{\small(a) \emph{room-32-32-4}}
    \end{minipage}%
    \hfill
    \begin{minipage}[t]{0.96\textwidth}
        \centering
        \begin{minipage}[t]{0.1625\textwidth}
            \includegraphics[width=\linewidth,trim={1.1cm 0 0 1.1cm},clip]{figures/room-32-32-4/LCBS.png} \\
            \centering \scriptsize LCBS (ours)
        \end{minipage}
        \hfill
        \begin{minipage}[t]{0.1625\textwidth}
            \includegraphics[width=\linewidth,trim={1.1cm 0 0 1.1cm},clip]{figures/room-32-32-4/BBMOCBS-k_1.png} \\
            \centering \scriptsize BB-MO-CBS-k (k=1)
        \end{minipage}
        \hfill
        \begin{minipage}[t]{0.1625\textwidth}
            \includegraphics[width=\linewidth,trim={1.1cm 0 0 1.1cm},clip]{figures/room-32-32-4/BBMOCBS-k_5.png} \\
            \centering \scriptsize BB-MO-CBS-k (k=5)
        \end{minipage}
        \hfill
        \begin{minipage}[t]{0.1625\textwidth}
            \includegraphics[width=\linewidth,trim={1.1cm 0 0 1.1cm},clip]{figures/room-32-32-4/BBMOCBS-k_10.png} \\
            \centering \scriptsize BB-MO-CBS-k (k=10)
        \end{minipage}
        \hfill
        \begin{minipage}[t]{0.1625\textwidth}
            \includegraphics[width=\linewidth,trim={1.1cm 0 0 1.1cm},clip]{figures/room-32-32-4/BBMOCBS-pex.png} \\
            \centering \scriptsize BB-MO-CBS-pex
        \end{minipage}
        \hfill
        \begin{minipage}[t]{0.1625\textwidth}
            \includegraphics[width=\linewidth,trim={1.1cm 0 0 1.1cm},clip]{figures/room-32-32-4/BBMOCBS-eps.png}\\
            \centering \scriptsize BB-MO-CBS-$\epsilon$
        \end{minipage}
    \end{minipage}
\end{minipage}
\end{center}

\begin{center}
\begin{minipage}[t]{\textwidth}
     \begin{minipage}[t]{0.02\textwidth}
        \centering
        \rotatebox{90}{\small(b) \emph{room-64-64-8}}
    \end{minipage}%
    \hfill
    \begin{minipage}[t]{0.96\textwidth}
        \centering
        \begin{minipage}[t]{0.1625\textwidth}
            \includegraphics[width=\linewidth,trim={1.1cm 0 0 1.1cm},clip]{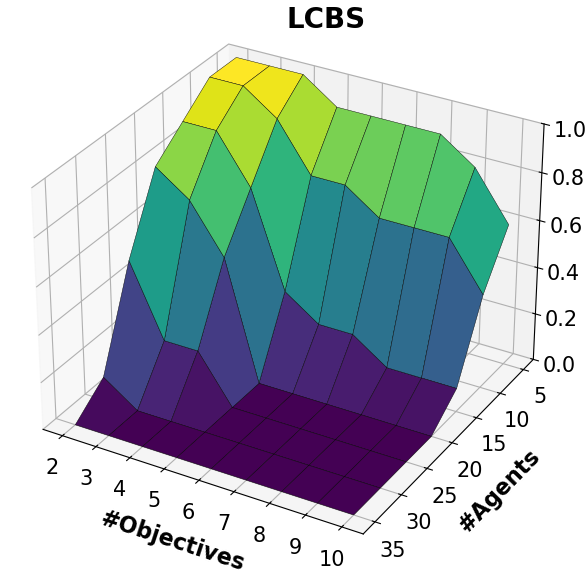} \\
            \centering \scriptsize LCBS (ours)
        \end{minipage}
        \hfill
        \begin{minipage}[t]{0.1625\textwidth}
            \includegraphics[width=\linewidth,trim={1.1cm 0 0 1.1cm},clip]{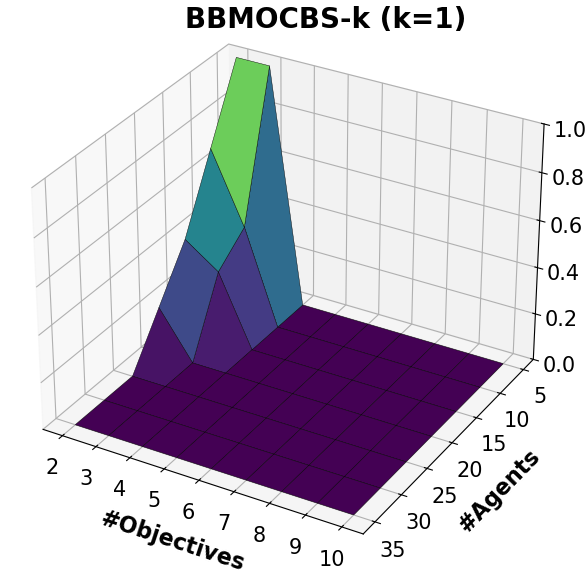} \\
            \centering \scriptsize BB-MO-CBS-k (k=1)
        \end{minipage}
        \hfill
        \begin{minipage}[t]{0.1625\textwidth}
            \includegraphics[width=\linewidth,trim={1.1cm 0 0 1.1cm},clip]{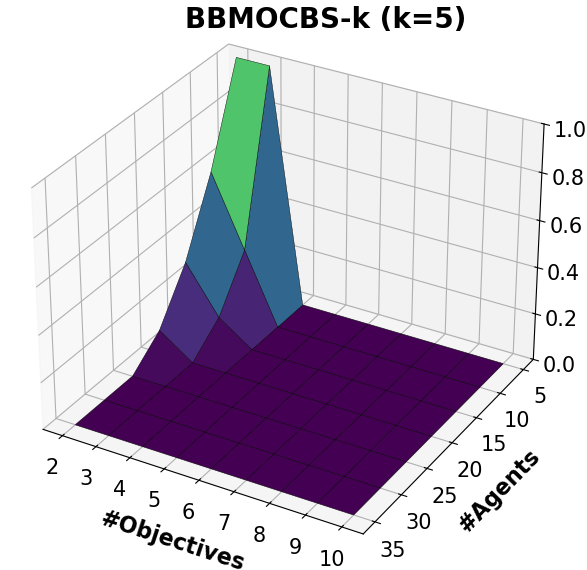} \\
            \centering \scriptsize BB-MO-CBS-k (k=5)
        \end{minipage}
        \hfill
        \begin{minipage}[t]{0.1625\textwidth}
            \includegraphics[width=\linewidth,trim={1.1cm 0 0 1.1cm},clip]{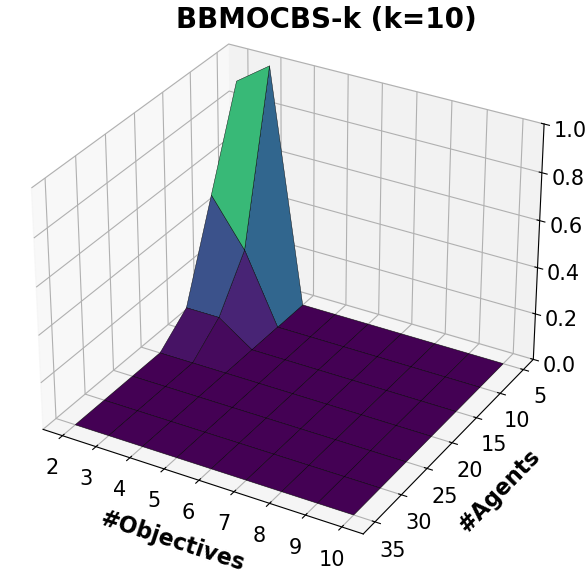} \\
            \centering \scriptsize BB-MO-CBS-k (k=10)
        \end{minipage}
        \hfill
        \begin{minipage}[t]{0.1625\textwidth}
            \includegraphics[width=\linewidth,trim={1.1cm 0 0 1.1cm},clip]{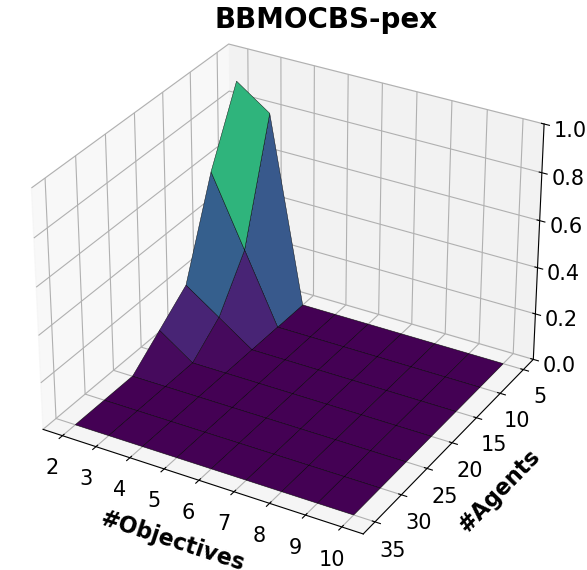} \\
            \centering \scriptsize BB-MO-CBS-pex
        \end{minipage}
        \hfill
        \begin{minipage}[t]{0.1625\textwidth}
            \includegraphics[width=\linewidth,trim={1.1cm 0 0 1.1cm},clip]{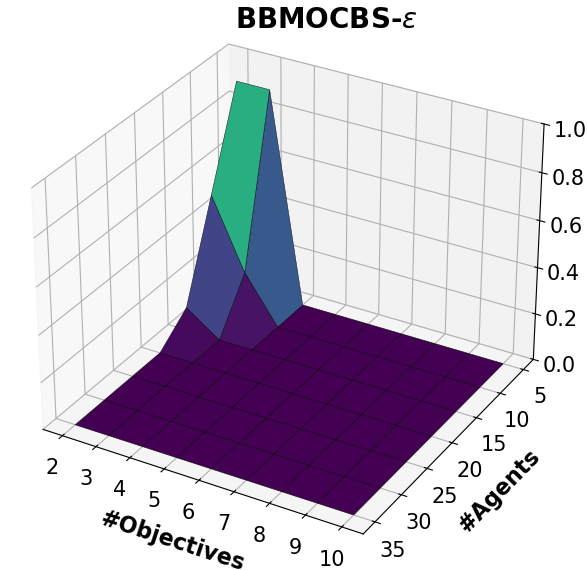}\\
            \centering \scriptsize BB-MO-CBS-$\epsilon$
        \end{minipage}
    \end{minipage}
\end{minipage}
\end{center}

\begin{center}
\begin{minipage}[t]{\textwidth}
     \begin{minipage}[t]{0.02\textwidth}
        \centering
        \rotatebox{90}{\small(c) \emph{maze-32-32-2}}
    \end{minipage}%
    \hfill
    \begin{minipage}[t]{0.96\textwidth}
        \centering
        \begin{minipage}[t]{0.1625\textwidth}
            \includegraphics[width=\linewidth,trim={1.1cm 0 0 1.1cm},clip]{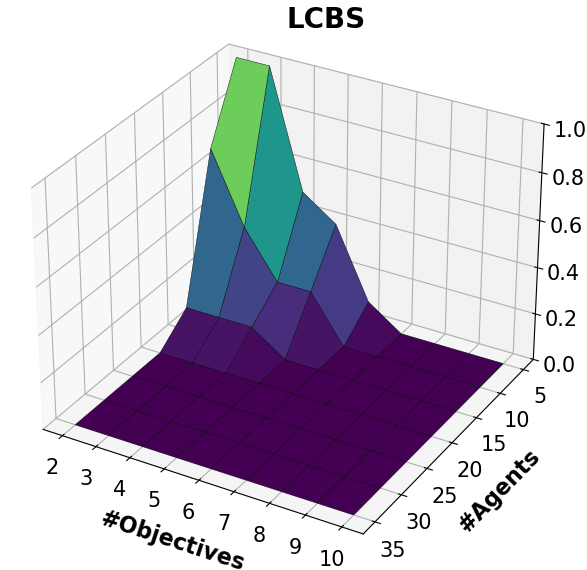} \\
            \centering \scriptsize LCBS (ours)
        \end{minipage}
        \hfill
        \begin{minipage}[t]{0.1625\textwidth}
            \includegraphics[width=\linewidth,trim={1.1cm 0 0 1.1cm},clip]{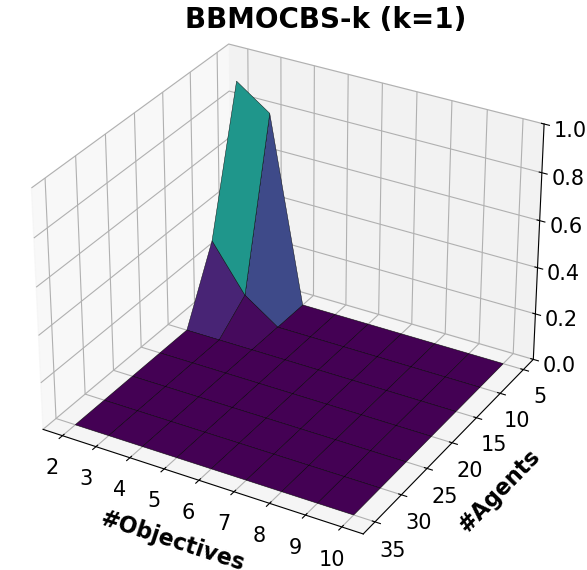} \\
            \centering \scriptsize BB-MO-CBS-k (k=1)
        \end{minipage}
        \hfill
        \begin{minipage}[t]{0.1625\textwidth}
            \includegraphics[width=\linewidth,trim={1.1cm 0 0 1.1cm},clip]{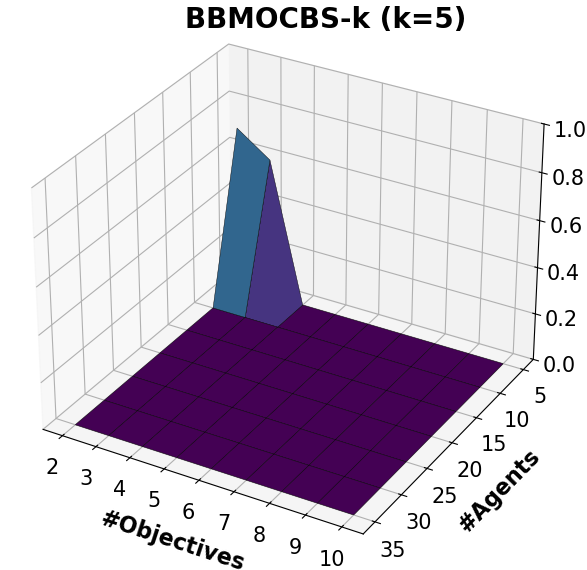} \\
            \centering \scriptsize BB-MO-CBS-k (k=5)
        \end{minipage}
        \hfill
        \begin{minipage}[t]{0.1625\textwidth}
            \includegraphics[width=\linewidth,trim={1.1cm 0 0 1.1cm},clip]{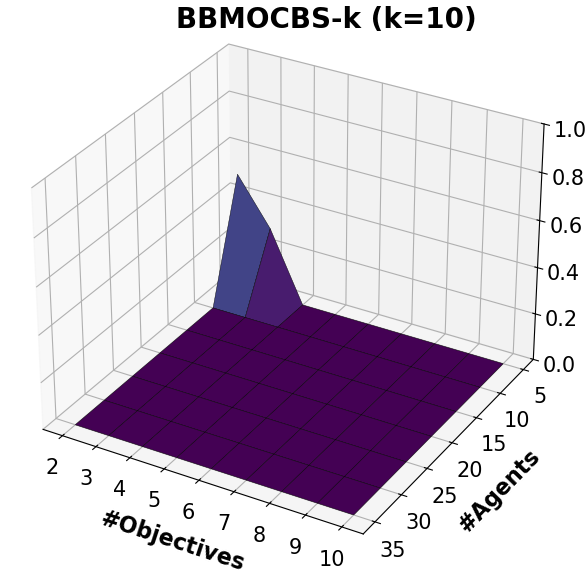} \\
            \centering \scriptsize BB-MO-CBS-k (k=10)
        \end{minipage}
        \hfill
        \begin{minipage}[t]{0.1625\textwidth}
            \includegraphics[width=\linewidth,trim={1.1cm 0 0 1.1cm},clip]{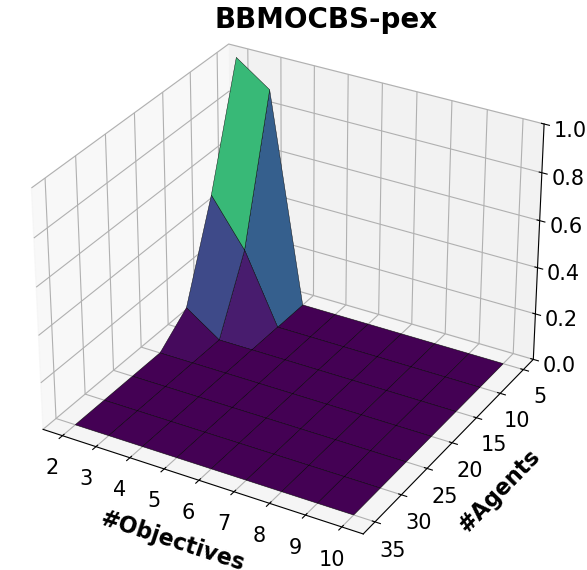} \\
            \centering \scriptsize BB-MO-CBS-pex
        \end{minipage}
        \hfill
        \begin{minipage}[t]{0.1625\textwidth}
            \includegraphics[width=\linewidth,trim={1.1cm 0 0 1.1cm},clip]{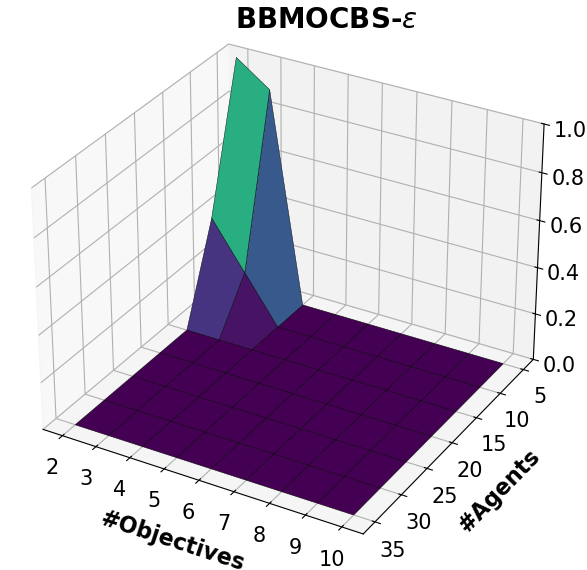}\\
            \centering \scriptsize BB-MO-CBS-$\epsilon$
        \end{minipage}
    \end{minipage}
\end{minipage}
\end{center}

\begin{center}
\begin{minipage}[t]{\textwidth}
     \begin{minipage}[t]{0.02\textwidth}
        \centering
        \rotatebox{90}{\small(d) \emph{maze-32-32-4}}
    \end{minipage}%
    \hfill
    \begin{minipage}[t]{0.96\textwidth}
        \centering
        \begin{minipage}[t]{0.1625\textwidth}
            \includegraphics[width=\linewidth,trim={1.1cm 0 0 1.1cm},clip]{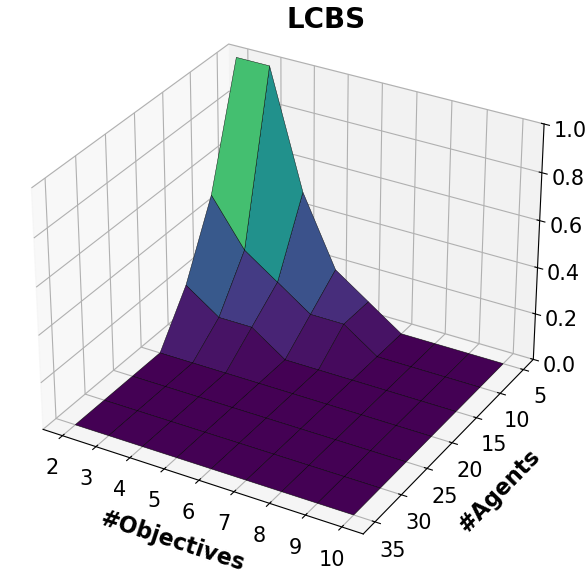} \\
            \centering \scriptsize LCBS (ours)
        \end{minipage}
        \hfill
        \begin{minipage}[t]{0.1625\textwidth}
            \includegraphics[width=\linewidth,trim={1.1cm 0 0 1.1cm},clip]{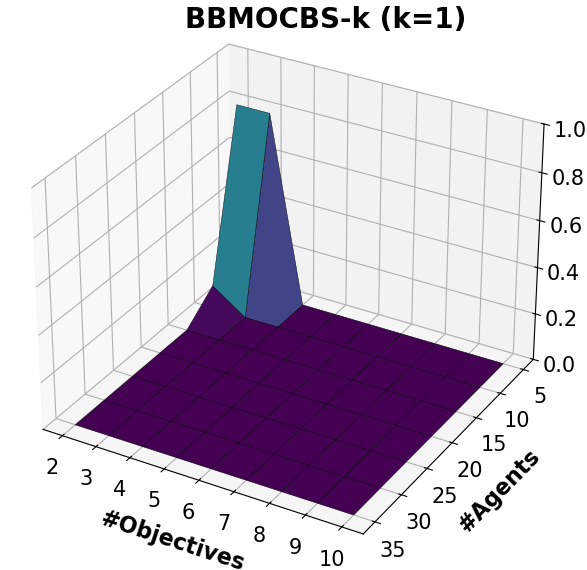} \\
            \centering \scriptsize BB-MO-CBS-k (k=1)
        \end{minipage}
        \hfill
        \begin{minipage}[t]{0.1625\textwidth}
            \includegraphics[width=\linewidth,trim={1.1cm 0 0 1.1cm},clip]{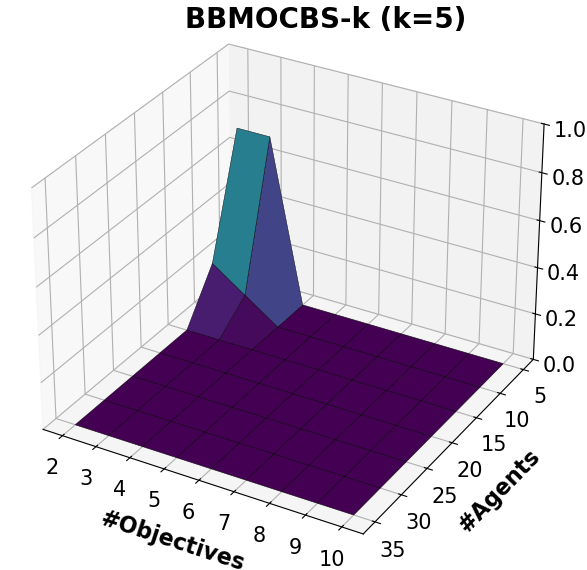} \\
            \centering \scriptsize BB-MO-CBS-k (k=5)
        \end{minipage}
        \hfill
        \begin{minipage}[t]{0.1625\textwidth}
            \includegraphics[width=\linewidth,trim={1.1cm 0 0 1.1cm},clip]{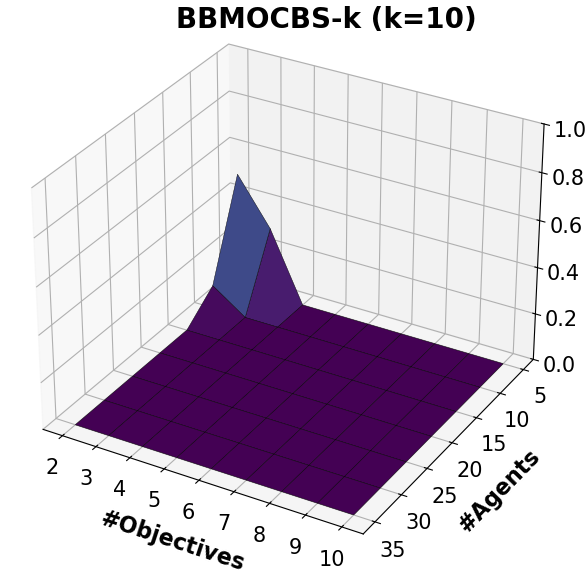} \\
            \centering \scriptsize BB-MO-CBS-k (k=10)
        \end{minipage}
        \hfill
        \begin{minipage}[t]{0.1625\textwidth}
            \includegraphics[width=\linewidth,trim={1.1cm 0 0 1.1cm},clip]{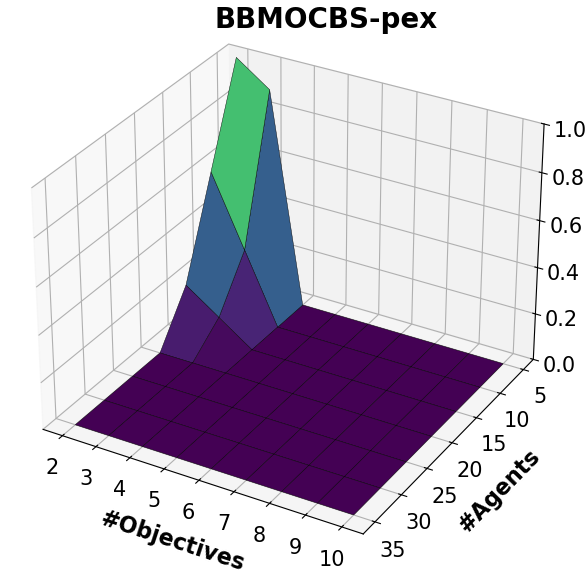} \\
            \centering \scriptsize BB-MO-CBS-pex
        \end{minipage}
        \hfill
        \begin{minipage}[t]{0.1625\textwidth}
            \includegraphics[width=\linewidth,trim={1.1cm 0 0 1.1cm},clip]{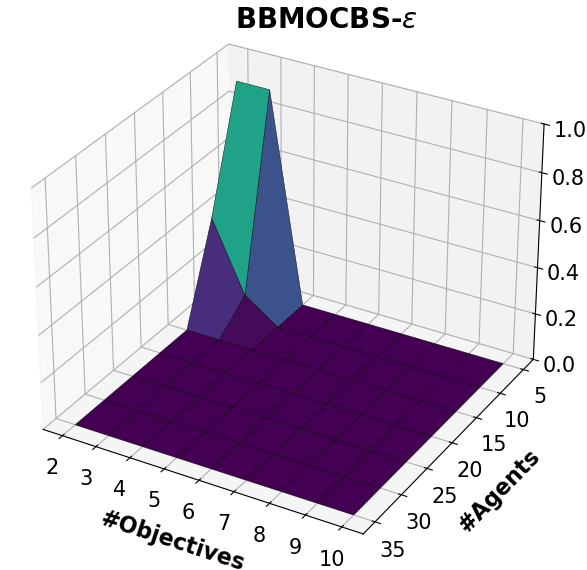}\\
            \centering \scriptsize BB-MO-CBS-$\epsilon$
        \end{minipage}
    \end{minipage}  \vspace{1ex}
    \centering
    
    {\vspace{5pt}\normalsize Figure 2: Success contours for MAPF benchmarks with 25 standard scenarios each ($T=5$ min)}
\end{minipage}
\end{center}

\vspace{2ex}
\begin{center}
\renewcommand{\arraystretch}{1.1}
\setlength{\tabcolsep}{4pt}

\begin{tabular}{|>{\centering\arraybackslash}p{0.18\textwidth}|
                >{\centering\arraybackslash}p{0.13\textwidth}|
                >{\centering\arraybackslash}p{0.13\textwidth}|
                >{\centering\arraybackslash}p{0.13\textwidth}|
                >{\centering\arraybackslash}p{0.15\textwidth}|
                >{\centering\arraybackslash}p{0.13\textwidth}|}
\hline
\multirow{2}{*}{\textbf{Approach}} 
& \multicolumn{5}{c|}{\textbf{Cost Vectors on Benchmarks with 5 Agents and 2 Objectives}} \\ \cline{2-6}
& \textbf{empty-32-32} & \textbf{maze-32-32-2} & \textbf{room-32-32-4} & \textbf{random-32-32-20} & \textbf{room-64-64-8} \\
\specialrule{1.1pt}{0pt}{0pt}
LCBS             & \{132, 246\} & \{96, 211\}  & \{139, 178\} & \{203, 216\} & \{117, 89\} \\ \hline
BBMOCBS-k (k=1)  & \{132, 246\} & \{96, 211\}  & \{139, 178\} & \{203, 216\} & \{117, 89\} \\ \hline
BBMOCBS-k (k=5)  & \{132, 246\} & \{96, 211\}  & \{139, 178\} & \{203, 216\} & \{117, 89\} \\ \hline
BBMOCBS-k (k=10) & \{132, 246\} & \{96, 211\}  & \{139, 178\} & \{203, 216\} & \{117, 89\} \\ \hline
BBMOCBS-pex      & \{132, 246\} & \{96, 211\}  & \{139, 178\} & \{203, 216\} & \{117, 89\} \\ \hline
BBMOCBS-$\epsilon$ & \{132, 246\} & \{96, 211\}  & \{139, 178\} & \{203, 216\} & \{117, 89\} \\ \hline
\end{tabular}

\vspace{2ex}

\begin{tabular}{|>{\centering\arraybackslash}p{0.18\textwidth}|
                >{\centering\arraybackslash}p{0.18\textwidth}|
                >{\centering\arraybackslash}p{0.175\textwidth}|
                >{\centering\arraybackslash}p{0.17\textwidth}|
                >{\centering\arraybackslash}p{0.17\textwidth}|}
\hline
\multirow{2}{*}{\textbf{Approach}} 
& \multicolumn{4}{c|}{\textbf{Cost Vectors on Benchmarks with 5 Agents and 2 Objectives}} \\ \cline{2-5}
& \textbf{random-64-64-10} & \textbf{random-64-64-20} & \textbf{maze-32-32-4} & \textbf{empty-48-48} \\
\specialrule{1.1pt}{0pt}{0pt}
LCBS             & \{198, 241\} & \{223, 268\} & \{113, 206\} & \{154, 221\} \\ \hline
BBMOCBS-k (k=1)  & \{198, 241\} & \{223, 268\} & \{113, 206\} & \{154, 221\} \\ \hline
BBMOCBS-k (k=5)  & \{198, 241\} & \{223, 268\} & \{113, 206\} & \{154, 221\} \\ \hline
BBMOCBS-k (k=10) & \{198, 241\} & \{223, 268\} & \{113, 206\} & \{154, 221\} \\ \hline
BBMOCBS-pex      & \{198, 241\} & \{223, 268\} & \{113, 206\} & \{154, 221\} \\ \hline
BBMOCBS-$\epsilon$ & \{198, 241\} & \{223, 268\} & \{113, 206\} & \{154, 221\} \\ \hline
\end{tabular}
\begin{minipage}{\textwidth}
    \centering
    \vspace{5pt}
    Table 1: Cost vectors from all approaches in selected scenarios across MAPF benchmarks with 5 agents and 2 objectives.
\end{minipage}
\end{center}


\begin{table*}[t]
\renewcommand{\arraystretch}{1.1}
\centering

\begin{tabular}{|>{\centering\arraybackslash}p{0.18\textwidth}|
                >{\centering\arraybackslash}p{0.13\textwidth}|
                >{\centering\arraybackslash}p{0.13\textwidth}|
                >{\centering\arraybackslash}p{0.13\textwidth}|
                >{\centering\arraybackslash}p{0.15\textwidth}|
                >{\centering\arraybackslash}p{0.13\textwidth}|}
\hline
\multirow{2}{*}{\textbf{Approach}} 
& \multicolumn{5}{c|}{\textbf{Cost Vectors on Benchmarks with 10 Agents and 2 Objectives}} \\ \cline{2-6}
& \textbf{empty-32-32} & \textbf{maze-32-32-2} & \textbf{room-32-32-4} & \textbf{random-32-32-20} & \textbf{room-64-64-8} \\
\specialrule{1.1pt}{0pt}{0pt}
LCBS             & \{284, 412\} & \{188, 369\}  & \{301, 342\} & \{392, 418\} & \{251, 177\} \\ \hline
BBMOCBS-k (k=1)  & \{284, 412\} & \{188, 369\}  & \{301, 342\} & \{392, 418\} & \{251, 177\} \\ \hline
BBMOCBS-k (k=5)  & \{284, 412\} & \{188, 369\}  & \{301, 342\} & \{392, 418\} & \{251, 177\} \\ \hline
BBMOCBS-k (k=10) & \{284, 412\} & \{188, 369\}  & \{301, 342\} & \{392, 418\} & \{251, 177\} \\ \hline
BBMOCBS-pex      & \{284, 412\} & \{188, 369\}  & \{301, 342\} & \{392, 418\} & \{251, 177\} \\ \hline
BBMOCBS-$\epsilon$ & \{284, 412\} & \{188, 369\}  & \{301, 342\} & \{392, 418\} & \{251, 177\} \\ \hline
\end{tabular}

\vspace{1.5ex}

\begin{tabular}{|>{\centering\arraybackslash}p{0.18\textwidth}|
                >{\centering\arraybackslash}p{0.18\textwidth}|
                >{\centering\arraybackslash}p{0.175\textwidth}|
                >{\centering\arraybackslash}p{0.17\textwidth}|
                >{\centering\arraybackslash}p{0.17\textwidth}|}
\hline
\multirow{2}{*}{\textbf{Approach}} 
& \multicolumn{4}{c|}{\textbf{Cost Vectors on Benchmarks with 10 Agents and 2 Objectives}} \\ \cline{2-5}
& \textbf{random-64-64-10} & \textbf{random-64-64-20} & \textbf{maze-32-32-4} & \textbf{empty-48-48} \\
\specialrule{1.1pt}{0pt}{0pt}
LCBS             & \{423, 467\} & \{448, 502\} & \{221, 396\} & \{318, 438\} \\ \hline
BBMOCBS-k (k=1)  & \{423, 467\} & \{448, 502\} & \{221, 396\} & \{318, 438\} \\ \hline
BBMOCBS-k (k=5)  & \{423, 467\} & \{448, 502\} & \{221, 396\} & \{318, 438\} \\ \hline
BBMOCBS-k (k=10) & \{423, 467\} & \{448, 502\} & \{221, 396\} & \{318, 438\} \\ \hline
BBMOCBS-pex      & \{423, 467\} & \{448, 502\} & \{221, 396\} & \{318, 438\} \\ \hline
BBMOCBS-$\epsilon$ & \{423, 467\} & \{448, 502\} & \{221, 396\} & \{318, 438\} \\ \hline
\end{tabular}

\caption{Cost vectors from all approaches in selected scenarios across MAPF benchmarks with 10 agents and 2 objectives.}
\label{tab:2obj-costs-10agents}
\end{table*}

\begin{table*}[t]
\renewcommand{\arraystretch}{1.1}
\centering

\begin{tabular}{|>{\centering\arraybackslash}p{0.18\textwidth}|
                >{\centering\arraybackslash}p{0.13\textwidth}|
                >{\centering\arraybackslash}p{0.13\textwidth}|
                >{\centering\arraybackslash}p{0.13\textwidth}|
                >{\centering\arraybackslash}p{0.15\textwidth}|
                >{\centering\arraybackslash}p{0.13\textwidth}|}
\hline
\multirow{2}{*}{\textbf{Approach}} 
& \multicolumn{5}{c|}{\textbf{Cost Vectors on Benchmarks with 5 Agents and 3 Objectives}} \\ \cline{2-6}
& \textbf{empty-32-32} & \textbf{maze-32-32-2} & \textbf{room-32-32-4} & \textbf{random-32-32-20} & \textbf{room-64-64-8} \\
\specialrule{1.1pt}{0pt}{0pt}
LCBS             & \{137, 268, 182\} & \{86, 188, 237\}  & \{149, 209, 157\} & \{221, 239, 229\} & \{132, 98, 226\} \\ \hline
BBMOCBS-k (k=1)  & \{137, 268, 182\} & \{86, 188, 237\}  & \{149, 209, 157\} & \{221, 239, 229\} & \{132, 98, 226\} \\ \hline
BBMOCBS-k (k=5)  & \{137, 268, 182\} & \{86, 188, 237\}  & \{149, 209, 157\} & \{221, 239, 229\} & \{132, 98, 226\} \\ \hline
BBMOCBS-k (k=10) & \{137, 268, 182\} & \{86, 188, 237\}  & \{149, 209, 157\} & \{221, 239, 229\} & \{132, 98, 226\} \\ \hline
BBMOCBS-pex      & \{137, 268, 182\} & \{86, 188, 237\}  & \{149, 209, 157\} & \{221, 239, 229\} & \{132, 98, 226\} \\ \hline
BBMOCBS-$\epsilon$ & \{137, 268, 182\} & \{86, 188, 237\}  & \{149, 209, 157\} & \{221, 239, 229\} & \{132, 98, 226\} \\ \hline
\end{tabular}

\vspace{1.5ex}

\begin{tabular}{|>{\centering\arraybackslash}p{0.18\textwidth}|
                >{\centering\arraybackslash}p{0.18\textwidth}|
                >{\centering\arraybackslash}p{0.175\textwidth}|
                >{\centering\arraybackslash}p{0.17\textwidth}|
                >{\centering\arraybackslash}p{0.17\textwidth}|}
\hline
\multirow{2}{*}{\textbf{Approach}} 
& \multicolumn{4}{c|}{\textbf{Cost Vectors on Benchmarks with 5 Agents and 3 Objectives}} \\ \cline{2-5}
& \textbf{random-64-64-10} & \textbf{random-64-64-20} & \textbf{maze-32-32-4} & \textbf{empty-48-48} \\
\specialrule{1.1pt}{0pt}{0pt}
LCBS             & \{201, 267, 298\} & \{224, 289, 276\} & \{106, 188, 217\} & \{148, 261, 183\} \\ \hline
BBMOCBS-k (k=1)  & \{201, 267, 298\} & \{224, 289, 276\} & \{106, 188, 217\} & \{148, 261, 183\} \\ \hline
BBMOCBS-k (k=5)  & \{201, 267, 298\} & \{224, 289, 276\} & \{106, 188, 217\} & \{148, 261, 183\} \\ \hline
BBMOCBS-k (k=10) & \{201, 267, 298\} & \{224, 289, 276\} & \{106, 188, 217\} & \{148, 261, 183\} \\ \hline
BBMOCBS-pex      & \{201, 267, 298\} & \{224, 289, 276\} & \{106, 188, 217\} & \{148, 261, 183\} \\ \hline
BBMOCBS-$\epsilon$ & \{201, 267, 298\} & \{224, 289, 276\} & \{106, 188, 217\} & \{148, 261, 183\} \\ \hline
\end{tabular}

\caption{Cost vectors from all approaches in selected scenarios across MAPF benchmarks with 5 agents and 3 objectives.}
\label{tab:optimality-comparison-additional}
\end{table*}

\begin{table*}[t]
\renewcommand{\arraystretch}{1.1}
\centering

\begin{tabular}{|>{\centering\arraybackslash}p{0.18\textwidth}|
                >{\centering\arraybackslash}p{0.13\textwidth}|
                >{\centering\arraybackslash}p{0.13\textwidth}|
                >{\centering\arraybackslash}p{0.13\textwidth}|
                >{\centering\arraybackslash}p{0.15\textwidth}|
                >{\centering\arraybackslash}p{0.13\textwidth}|}
\hline
\multirow{2}{*}{\textbf{Approach}} 
& \multicolumn{5}{c|}{\textbf{Cost Vectors on Benchmarks with 10 Agents and 3 Objectives}} \\ \cline{2-6}
& \textbf{empty-32-32} & \textbf{maze-32-32-2} & \textbf{room-32-32-4} & \textbf{random-32-32-20} & \textbf{room-64-64-8} \\
\specialrule{1.1pt}{0pt}{0pt}
LCBS             & \{247, 493, 343\} & \{160, 358, 456\} & \{284, 408, 315\} & \{408, 452, 436\} & \{254, 185, 388\} \\ \hline
BBMOCBS-k (k=1)  & \{247, 493, 343\} & \{160, 358, 456\} & \{284, 408, 315\} & \{408, 452, 436\} & \{254, 185, 388\} \\ \hline
BBMOCBS-k (k=5)  & \{247, 493, 343\} & \{160, 358, 456\} & \{284, 408, 315\} & \{408, 452, 436\} & \{254, 185, 388\} \\ \hline
BBMOCBS-k (k=10) & \{247, 493, 343\} & \{160, 358, 456\} & \{284, 408, 315\} & \{408, 452, 436\} & \{254, 185, 388\} \\ \hline
BBMOCBS-pex      & \{247, 493, 343\} & \{160, 358, 456\} & \{284, 408, 315\} & \{408, 452, 436\} & \{254, 185, 388\} \\ \hline
BBMOCBS-$\epsilon$ & \{247, 493, 343\} & \{160, 358, 456\} & \{284, 408, 315\} & \{408, 452, 436\} & \{254, 185, 388\} \\ \hline
\end{tabular}
\vspace{1.5ex}

\begin{tabular}{|>{\centering\arraybackslash}p{0.18\textwidth}|
                >{\centering\arraybackslash}p{0.18\textwidth}|
                >{\centering\arraybackslash}p{0.175\textwidth}|
                >{\centering\arraybackslash}p{0.17\textwidth}|
                >{\centering\arraybackslash}p{0.17\textwidth}|}
\hline
\multirow{2}{*}{\textbf{Approach}} 
& \multicolumn{4}{c|}{\textbf{Cost Vectors on Benchmarks with 10 Agents and 3 Objectives}} \\ \cline{2-5}
& \textbf{random-64-64-10} & \textbf{random-64-64-20} & \textbf{maze-32-32-4} & \textbf{empty-48-48} \\
\specialrule{1.1pt}{0pt}{0pt}
LCBS             & \{362, 481, 543\} & \{419, 558, 497\} & \{190, 358, 420\} & \{296, 435, 341\} \\ \hline
BBMOCBS-k (k=1)  & \{362, 481, 543\} & \{419, 558, 497\} & \{190, 358, 420\} & \{296, 435, 341\} \\ \hline
BBMOCBS-k (k=5)  & \{362, 481, 543\} & \{419, 558, 497\} & \{190, 358, 420\} & \{296, 435, 341\} \\ \hline
BBMOCBS-k (k=10) & \{362, 481, 543\} & \{419, 558, 497\} & \{190, 358, 420\} & \{296, 435, 341\} \\ \hline
BBMOCBS-pex      & \{362, 481, 543\} & \{419, 558, 497\} & \{190, 358, 420\} & \{296, 435, 341\} \\ \hline
BBMOCBS-$\epsilon$ & \{362, 481, 543\} & \{419, 558, 497\} & \{190, 358, 420\} & \{296, 435, 341\} \\ \hline
\end{tabular}

\caption{Cost vectors from all approaches in selected scenarios across MAPF benchmarks with 10 agents and 3 objectives.}
\label{tab:optimality-comparison-10agents}
\end{table*}
\hfill